\documentclass[pdflatex,sn-mathphys-num]{sn-jnl}

\usepackage{graphicx}%
\usepackage{multirow}%
\usepackage{amsmath,amssymb,amsfonts}%
\usepackage{amsthm}%
\usepackage{mathrsfs}%
\usepackage[title]{appendix}%
\usepackage{xcolor}%
\usepackage{textcomp}%
\usepackage{manyfoot}%
\usepackage{booktabs}%
\usepackage{algorithm}%
\usepackage{algorithmicx}%
\usepackage{algpseudocode}%
\usepackage{listings}%
\usepackage{adjustbox}
\usepackage{enumitem}
\usepackage{amsfonts}
\usepackage[utf8]{inputenc}

\theoremstyle{thmstyleone}%
\theoremstyle{thmstyletwo}%

\theoremstyle{thmstylethree}%

\begin{document}

\title[Article Title]{TabulaTime: A Novel Multimodal Deep Learning Framework for
Advancing Acute Coronary Syndrome Prediction through Environmental and
Clinical Data Integration}


\author[1]{\fnm{Xin} \sur{Zhang}}\email{x.zhang@mmu.ac.uk}

\author*[1]{\fnm{Liangxiu} \sur{Han}}\email{l.han@mmu.ac.uk}

\author[2]{\fnm{Saad} \sur{Hassan}}\email{saad.hassan@doctors.org.uk}
\author[3]{\fnm{Philip} \sur{A Kalra}}\email{philip.kalra@nca.nhs.uk}
\author[3]{\fnm{James} \sur{Ritchie}}\email{james.ritchie@nca.nhs.uk}

\author[4]{\fnm{Carl} \sur{Diver}}\email{C.Diver@mmu.ac.uk}

\author[5]{\fnm{Jennie} \sur{Shorley}}\email{j.shorley@mmu.ac.uk}
\author*[6]{\fnm{Stephen} \sur{White}}\email{steve.white3@newcastle.ac.uk}

\affil*[1]{Department of Computing, and Mathematics, Manchester Metropolitan
University, Manchester, M15 6BH, UK}

\affil[2]{Blackpool Teaching Hospitals NHS Foundation Trust, Blackpool Victoria Hospital, Whinney Heys Road Blackpool, FY3 8NR}

\affil[3]{Northern Care Alliance, NHS Foundation Trust, Salford, M6 8HD, UK}
\affil[4]{Department of Engineering, Manchester Metropolitan University,Manchester, M15 6BH, UK}
\affil[5]{Faculty of Business and Law, Manchester Metropolitan University,Manchester, M15 6BH, UK}
\affil*[6]{Faculty of Medical Sciences, Newcastle University, Newcastle upon
Tyne, NE1 3BZ, UK}

\abstract{

Acute Coronary Syndromes (ACS), encompassing conditions such as ST-segment elevation myocardial infarction (STEMI) and non-ST-segment elevation myocardial infarction (NSTEMI), remain a leading cause of global mortality. While traditional Cardiovascular Risk Scores (CVRS) provide valuable insights, they primarily rely on clinical data, often overlooking environmental factors like air pollution, which significantly impact cardiovascular health. Additionally, integrating complex time-series environmental data with clinical datasets poses significant challenges due to issues in alignment and fusion.

To address these gaps, we propose TabulaTime, a novel multimodal deep learning framework that integrates clinical risk factors with air pollution and climate data to enhance ACS risk prediction. TabulaTime introduces three major innovations: (1) Multimodal feature integration to combine time-series environmental data (air pollution and climate) with clinical tabular data for improved predictive accuracy; (2) PatchRWKV, a new module for automatic extraction of complex temporal patterns, overcoming limitations of traditional feature engineering and maintaining linear computational complexity; and (3) Enhanced interpretability using attention mechanisms to reveal interactions between clinical and environmental factors.

Our experimental evaluation demonstrated that TabulaTime achieves a 20.5\% improvement in accuracy compared to the best-performing traditional machine learning model (CatBoost) and surpasses other models, including Random Forest and LightGBM, by 20.5\% to 32.2\% in differentiating ACS presentation (STEMI vs. NSTEMI). The integration of air pollution and climate data improves accuracy by 10.1\%, emphasising the importance of environmental factors. The PatchRWKV module showed advanced performance in time-series analysis, outperforming state-of-the-art methods such as Multi-Layer Perceptrons (MLPs) based methods (LightTS and DLinear), Convolutional Neural Networks (CNNs) based methods (TimesNet), Recurrent Neural Networks (RNNs) based methods (LSTM, GRU), and Transformers based methods (Autoformer, PatchTST) across various classification and forecasting tasks while maintaining linear computational efficiency. Additionally, feature importance analysis identified key predictors, including clinical factors like Systolic Blood Pressure, Symptom-to-Admission Time, and BMI, alongside environmental factors such as PM$_{10}$ (maximum and average), average NO, and temperature, highlighting the critical role of environmental-clinical interactions in ACS presentation.

This framework bridges the gap between traditional clinical models and environmental health insights, supporting personalised prevention strategies and informing public health policies to mitigate the cardiovascular impact of air pollution.
}




\maketitle

\section{Introduction}

Acute Coronary Syndromes (ACS), a spectrum of conditions encompassing unstable angina, ST-segment elevation myocardial infarction (STEMI), and non-ST-segment elevation myocardial infarction (NSTEMI), is the leading cause of death globally, with nearly half of these deaths due to ischaemic heart disease \cite{Reed2017Acute,Bergmark2022Acute}.  Understanding environmental factors is crucial for developing strategies to mitigate ACS risk. Cardiovascular Risk Scores (CVRS) serve as statistical tools used to estimate a person’s risk of having a cardiovascular event, such as a heart attack or stroke, usually over the next 10 years. Various risk scores have been developed, each targeting different populations, incorporating distinct risk factors, and possessing unique predictive strengths. Notable examples include the QRISK family of algorithms \cite{Hippisley2017Development,Hippisley2007Derivation,Hippisley2010Derivation}, Framingham Risk Score (FRS)\cite{Hemann2007Framingham}, Reynolds Risk Score \cite{Ridker2007Development} and European SCORE System \cite{Nashef1999European}. These tools help clinicians assess and manage patients’ cardiovascular risk and target prophylactic therapy more effectively.

However, there is a growing recognition that existing models may not fully capture the comprehensive range of risk factors contributing to ACS. For example, QRISK3 explains 59.6\% of the variation in time to diagnosis of CVD among women and 54.8\% among men, leaving over 40\% of the variation unexplained by the QRISK3 score \cite{Hippisley2017Development}. Numerous epidemiological studies have convincingly linked air pollution exposure to a heightened risk of ACS \cite{Rus2022Impact,Chen2022Hourly}. Air pollutants, particularly particulate matter (PM$_{2.5}$ and PM$_{10}$), nitrogen oxides (NO$_x$), sulphur dioxide (SO$_{2}$), and ozone (O$_3$), trigger inflammation, oxidative stress, and change in blood vessel function, contributing to cardiovascular issues. For instance,  \citeauthor{Rus2022Impact} \cite{Rus2022Impact}provides a broader look at the connection between environmental factors and ACS, highlighting the increased risk associated with PM$_{2.5}$ exposure. Additionally, in \cite{Chen2022Hourly}, the researchers presented a large-scale study analysing the link between hourly air pollutant levels and ACS in over 1 million patients, emphasising the roles of PM$_{2.5}$, NO$_{2}$, SO$_{2}$, and CO in triggering ACS. This gap necessitates novel approaches that leverage the power of environmental data to enhance risk prediction accuracy. 

Traditional medical data used for acute coronary syndrome (ACS) prediction typically include patient demographics and clinical data. These data points, structured in a tabular manner, allow for relatively simple analysis and prediction models. In contrast, the environmental data exists as a time series with more complex features, capturing continuous changes over time. Most current studies often utilise manual time series features, such as statistical features (e.g., maximum and minimum values, standard deviation) and temporal features (e.g., trends, event counts, anomalies). However, these methods do not fully leverage the extensive information embedded within time series data. Identifying and extracting the right features that represent the temporal dependencies within time-series data can be time-consuming and requires domain expertise \cite{Torres2021Deep}.

Deep learning models, such as recurrent neural networks (RNNs) \cite{Connor1994Recurrent}, multi-layer perceptrons (MLPs) \cite{Mayer1999Evolutionary}, convolutional neural networks (CNNs) \cite{Zhao2017Convolutional} and transformers \cite{Vaswani2017Attention}, can automatically learn complex temporal features from raw time series data, reducing the need for extensive manual engineering. Several studies have leveraged deep learning to analyse the relationship between air pollution exposure and health outcomes \cite{Gugnani2022Analysis,Zhou2020Deep,Ghufran2021Forecast,Subramaniam2022Artificial}. For instance, \cite{Bekkar2021Air} implemented a hybrid CNN-LSTM (Long Short-Term Memory) model to analyse hourly PM$_{2.5}$ concentrations in Beijing, China, and predict air quality levels for health risk assessment. Similarly, (Tsai, Zeng, and Chang 2018) used RNNs with LSTM to forecast PM$_{2.5}$ concentrations, while\cite{Miranda2021Application} employed radial basis function (RBF) and MLP neural networks to model the temporal dynamics of air pollution and its impact on respiratory hospital admissions.
Despite the success of advanced deep learning models in various fields, they can demonstrate limitations when processing multimodal data \cite{Jabeen2023Review} . Integrating heterogeneous data types requires sophisticated architectures capable of learning and representing the intricate relationships between different modalities. Furthermore, the alignment, synchronisation, and fusion of multimodal data necessitates advanced techniques to ensure that the combined information enhances, rather than detracts from, the model’s performance \cite{Gandhi2023Multimodal,Liang2024Foundations}.

Existing research on ACS risk prediction primarily emphasises traditional clinical datasets, often neglecting the significant influence of environmental factors. Integrating air pollution data into deep learning models offers a promising opportunity to enhance ACS risk assessment and inform preventive cardiology strategies. This research aims to address these challenges by developing and validating a novel multimodal deep learning based ACS prediction model, TabulaTime, specifically designed to improve CVD risk prediction through the integration of time-series air pollution features and clinical tabular data. Our proposed method offers several key contributions:

\begin{enumerate}
\item [1)]
\textbf{Multimodal feature integration and ACS prediction}: We propose a novel TabulaTime deep learning framework for multimodal feature integration and ACS prediction, combining extracted time-series features from environmental datasets with ACS risk factors obtained from clinical tabular data. This integration enables the model to conditionally learn attention maps based on the multimodal features, thereby enhancing both predictive performance and interpretability.

\item [2)]
\textbf{Automatic time-series feature extraction}: We introduce PatchRWKV, an advanced feature extraction module designed for automatic extraction of features from time-series data. The PatchRWKV module divides data into patches. It uses the Receptance Weighted Key Value (RWKV) technique, which combines RNN capabilities with attention mechanisms. This approach efficiently processes long sequences while maintaining linear computational complexity, making it more efficient than traditional RNNs and transformers, which have quadratic complexity. By capturing both short-term and long-term patterns, PatchRWKV provides a comprehensive representation of temporal dependencies. This capability is crucial for accurately detecting significant associations between pollution exposure patterns and ACS risk, which are often missed by traditional manual feature extraction methods.  

\item [3)]
\textbf{Enhanced interpretability}: We employ attention mechanisms and feature importance analysis to uncover potential interactions between air pollution, clinical risk factors, and the onset of ACS. This approach ensures that the model’s predictions are both accurate and interpretable, providing deeper insights into how environmental and clinical factors contribute to ACS risk. 

\end{enumerate}

This paper is structured as follows: Section 2 reviews related work, Section 3 details the proposed TabulaTime framework, and Sections 4 and 5 present experimental evaluations and discussions. The conclusion summarises findings and implications for future research.

\section{Related Work}

\subsection{Traditional ACS Prediction Methods}

Acute Coronary Syndrome (ACS) risk prediction methods are essential tools for estimating an individual's likelihood of experiencing cardiovascular events. Traditional methods for predicting ACS risk such as Framingham Risk Score (FRS)\cite{Hemann2007Framingham}, QRISK\cite{Hippisley2017Development}, Reynolds Risk Score \cite{Ridker2007Development} and European SCORE System \cite{Nashef1999European} utilise a combination of demographic, clinical, lifestyle and medical history data to provide an overall risk assessment, but frequently cannot be utilised once a diagnosis of underlying cardiovascular disease has been made. The data used can be categorised as follows: 1) Demographic Information: Age, gender, and family history of cardiovascular diseases\cite{Steen2022Event}. 2) Clinical History: Previous incidences of myocardial infarction, angina, hypertension, diabetes, and hypercholesterolemia. 3) Laboratory and and Clinical Tests: Blood pressure readings, cholesterol levels, and creatinine levels. 4) Electrocardiogram (ECG) Readings: Used to diagnose the type and severity of myocardial infarctions (e.g., Discrimination between STEMI and NSTEMI) \cite{Bhatt2022Diagnosis}.

These traditional methods systematically assess cardiovascular event risk. The FRS, Reynolds Risk Score, National Early Warning Score (NEWS) \cite{Smith2013ability}  and Global Registry of Acute Coronary Events (GRACE) (‘Rationale and Design of the GRACE (Global Registry of Acute Coronary Events) Project: A Multinational Registry of Patients Hospitalized with Acute Coronary Syndromes’ 2001) primarily relied on point-based systems incorporating traditional clinical and lifestyle factors. Each risk factor was assigned a specific number of points based on predefined scales. The QRISK algorithm employed a comprehensive multivariate approach using a wide array of data points to reflect a diverse population. The European SCORE system utilised regional risk charts to account for geographic variability in cardiovascular disease prevalence. 

\subsection{Machine Learning in ACS Prediction}

While these traditional models have been instrumental in guiding clinical practice, they have notable limitations in adaptability, data integration, and predictive accuracy.  Machine learning/deep learning methods offer significant advantages in these areas, providing more dynamic, accurate, and comprehensive risk assessments by leveraging advanced data analytics and continuous learning capabilities \cite{Ke2022Machine}.  Wu et al. \cite{Wu2021Machine} utilised machine learning to predict in-hospital cardiac arrest in ACS patients, finding that the XGBoost model outperformed traditional risk scores such as GRACE and NEWS , achieving high accuracy and AUC. A total of 45 risk features were selected in this work from the electronic health record including age, gender, history of smoking and laboratory features, Killip classification, vital signs, mental status, etc.  Similarly, Hadanny et al. \cite{Hadanny2022Machine} used Random Survival Forest (RSF) and deep neural network (DeepSurv) models to predict 1-year mortality in ACS patients, highlighting the improved performance of RSF over traditional methods. Acute Coronary Syndrome Israeli Survey (ACSIS) and the Myocardial Ischemia National Audit Project (MINAP) data were used in this work. 69 risk factors including demographics, prior medical history, prior medication, clinical presentation, basic laboratory data with admission were selected and evaluated.

D’Ascenzo et al. \cite{D2021Machine} developed the PRAISE risk scores, a machine learning tool validated with external cohorts, which showed high accuracy in predicting post-discharge outcomes for ACS patients. The 25 risk factors included 16 clinical variables, 5 therapeutic variables, and 2 angiographic variables. Research by Emakhu et al. \cite{Emakhu2022Acute} compared various machine learning techniques for predicting ACS outcomes based on 58 variables, including demographic and clinical factors ( e.g. brain natriuretic peptide (BNP), creatinine, glucose, heart rate, red cell distribution width, systolic blood pressure, and troponin), with models like XGBoost and RSF significantly improving prediction accuracy over traditional methods. Ke et al. \cite{Ke2022Machine} focused on early ACS onset prediction using ensemble methods such as gradient boosting, demonstrating the benefits of integrating machine learning into clinical practice for enhanced early detection and management of ACS using demographic characteristics, comorbidities, thrombolytic therapy, laboratory test data, and physical examination data. The study in \cite{Lee2021Machine} concluded that the ML-based approach improved the prediction of mortality, particularly in patients with non-ST-segment elevation myocardial infarction (NSTEMI) based on demographic characteristics, medical history, symptom, initial presentation, laboratory findings, clinical manifestation, echocardiographic finding, coronary angiographic finding, and medication at discharge. 

In recent years, deep learning techniques have shown significant advancements in various healthcare applications, including cardiovascular disease prediction. In \cite{duan2019utilizing}, a deep learning-based approach is introduced to analyse a massive volume of heterogeneous electronic health records in order to predict MACEs following ACS. In \cite{liu2023deep}and \cite{liu2021deep}, the authors introduce deep learning model for classifying ACS abnormalities using electrocardiogram (ECG) data. However, due to the structured and tabular nature of clinical data, the application of deep learning models in ACS prediction remains relatively limited compared to traditional machine learning approaches.

\subsection{Environmental Factors and ACS}

The effects of environmental factors on ACS have also been increasingly studied. Ku\'{z}ma et al. \cite{Kuzma2021Impact} investigated the short-term impact of air pollution on ACS incidence in industrial versus non-industrial areas, finding significant associations between air pollution and ACS admissions. Chen et al. \cite{Chen2022Hourly} examined hourly air pollutant concentrations and their association with ACS onset, employing a case-crossover design that revealed a strong link between short-term exposure to pollutants and increased ACS risk. Gestro et al. \cite{Gestro2020Short} developed models to analyse the delayed effects of air pollutants on emergency department admissions for ACS, underscoring the need for continuous air quality monitoring.

However, environmental time series data have more complex temporal properties than clinic data.  Current approaches often rely on manually selected features from time series datasets, including basic statistical measures such as mean, variance, and autocorrelation. These methods may not capture all relevant information, particularly complex, non-linear patterns. Recently, deep learning has emerged as a powerful tool for time series analysis, especially in examining the impact of air pollution on health outcomes. Various deep learning algorithms have been employed to analyse time series environmental data, such as Recurrent Neural Networks (RNNs) \cite{Husken2003Recurrent,Hewamalage2021Recurrent}, Multi-Layer Perceptrons (MLPs)\cite{Mayer1999Evolutionary}, Convolutional Neural Networks (CNNs)\cite{Zhao2017Convolutional}, and Transformers \cite{Zeng2023Are}.

\textbf{RNNs} are specifically designed for sequence data, making them a natural fit for time series analysis. Their internal state allows them to retain information from previous inputs, which is crucial for understanding temporal dependencies. For instance, Villegas et al. \cite{Villegas2023Predicting} proposed a predictive model for COVID-19 mortality risk using RNNs with attention mechanisms to enhance interpretability. Results indicate that the RNN model outperforms traditional baselines like Support Vector Classifier and Random Forest in sensitivity and overall stability.  

\textbf{MLPs}, though simpler than other models, can effectively model temporal dependencies when applied along the temporal dimension. MLP methods encode these temporal dependencies into the fixed parameters of MLP layers, adopting the MLP framework along the temporal axis. This approach allows MLPs to model sequential patterns in time-series data, providing a straightforward yet powerful means to analyse complex temporal relationships. Suttaket and Kok \cite{Suttaket2024Interpretable} introduced Rational Multi-Layer Perceptrons (RMLP) as a novel interpretable predictive model for healthcare, addressing the limitations of deep learning’s black-box nature. The model combined the strengths of weighted finite state automata and multi-layer perceptrons to process sequential data from electronic health records (EHRs). The study demonstrated that RMLP achieved strong predictive accuracy and enhanced interpretability across six clinical tasks. 

\textbf{CNNS} Convolutional Neural Networks (CNNs), commonly used for image processing, had also been successfully adapted for time-series analysis, demonstrating strength in capturing local patterns and hierarchical features through convolutional operations \cite{Muller2024IMU}. However, standard CNN architectures often struggled with modelling long-range temporal dependencies due to their limited receptive fields. To address this, several recent approaches in the Long-Term Time Series Forecasting (LTSF) domain introduced architectural innovations that extended temporal modelling capabilities while maintaining computational efficiency. For instance, SCINet \cite{liu2022scinet} employed a recursive downsample–convolve–interact mechanism to extract multi-resolution temporal features. MICN \cite{wang2023micn} used a multi-scale architecture that combined downsampled convolutions for local features with isometric convolutions for global context, achieving strong performance with linear complexity. TimesNet \cite{Wu2023TimesNet}, which we included as a CNN-based baseline in our evaluation, transformed 1D time series into 2D tensors and applied 2D convolutions to capture periodic and structured temporal variations. Despite these advancements, CNNs remained inherently limited in capturing extended temporal relationships—especially in clinical and environmental health contexts where delayed effects, such as those from air pollution exposure prior to ACS onset, were crucial. Extending CNNs' temporal range typically required deeper or dilated architectures, which increased model complexity and computational cost. These limitations motivated the development of more expressive and scalable frameworks capable of natively modelling long-range dependencies.

\textbf{Transformers}, introduced by Vaswani et al. \cite{Vaswani2017Attention}, have revolutionised sequence modelling by relying on self-attention mechanisms rather than recurrent structures. This allows for parallel processing of data and better handling of long-range dependencies. Ni et al. \cite{Ni2024Time} compared traditional time series models such as ARIMA and Prophet with advanced transformers based models for predicting heart rate dynamics. The study demonstrated that deep learning approaches, particularly transformer-based models like PatchTST \cite{Nie2023Time}, significantly outperformed traditional methods in capturing complex temporal dependencies and non-linear relationships. 

Despite their success, each of these models has specific strengths and limitations that make them suitable for different aspects of time series analysis. RNNs and LSTMs are excellent for handling temporal dependencies but are computationally intensive and complex. MLPs are simpler but less effective for complex time series compared to RNNs or LSTMs and less flexible for variable-length sequences due to requiring fixed input sizes. CNNs excel at capturing local patterns but often miss long-term dependencies. Transformers are powerful but have quadratic complexity relative to sequence length, making them computationally expensive for very long sequences.

Moreover, to the best of our knowledge, there are a few studies \cite{Sayed2024Novel} combining time series air pollution analysis and clinical data using deep learning algorithms for ACS prediction. Integrating heterogeneous data types remains challenging.  Multimodal deep learning frameworks are required to effectively combine clinical and environmental data, addressing issues related to alignment, synchronization, and fusion of multimodal information.  

Our proposed TabulaTime model represents a significant advancement in this field by leveraging deep learning to integrate time-series air pollution data with traditional clinical data. It efficiently processes long sequences while maintaining linear computational complexity, making it more efficient than traditional RNNs and transformers, which have quadratic complexity. This approach addresses the limitations of previous models, providing a more comprehensive and accurate assessment of ACS risk.

\section{The Proposed Method}

\subsection{Overview of the Framework}

\begin{figure}[!htbp]
\centering
\includegraphics[width=0.9\textwidth]{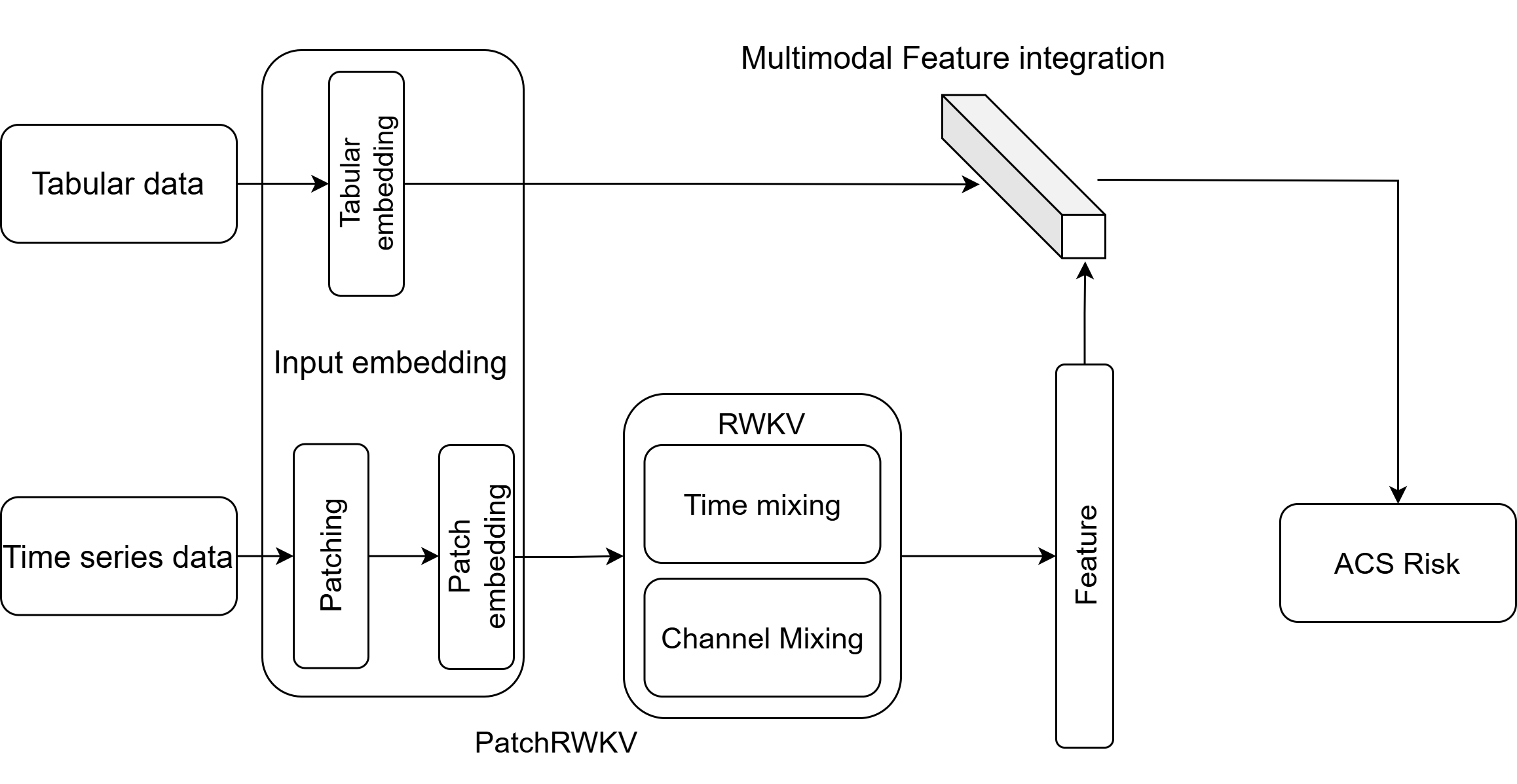}
\caption{The flowchart of the proposed TabulaTime framework.}
\label{fig:1}
\end{figure}

In this study, we propose a multimodal deep learning framework for modelling the effect of air pollution on ACS presentation by integrating time-series data with tabular data, named TabulaTime. {Figure~\ref{fig:1}} illustrates the flowchart of the TabulaTime framework, which comprises three main components: 

\begin{enumerate}
\item[1)]

Input embedding. This component converts raw input data, such as words, images, or time series data, into dense, continuous vectors that a deep learning model can process. This transformation enables the model to interpret and utilise the data more effectively. For instance, clinical tabular data are embedded into vectors, allowing the model to learn simultaneously from both clinical and other data types. This enhances the model’s ability to capture complex patterns and relationships, thereby improving its overall predictive performance.   
\item[2)]

Patched RWKV module (PatchRWKV) for automatic time-series sequential data feature extraction, which includes patching and RWKV (Receptance Weighted Key Value). The patching process divides time-series data into fixed-size segments, enabling the model to focus on short-term patterns within each segment. The RWKV module serves as the backbone for feature extraction, combining the strengths of recurrent neural networks (RNNs) and attention mechanisms to efficiently process time series sequences. PatchRWKV excels at identifying both short-term spikes and long-term trends in the data, providing a comprehensive representation of temporal dependencies. This approach efficiently processes long sequences while maintaining linear computational complexity, making it more efficient than traditional RNNs and transformers, which have quadratic complexity. By capturing both short-term and long-term patterns, PatchRWKV offers a thorough representation of temporal dependencies, crucial for accurately detecting significant associations between pollution exposure patterns and ACS risk, often missed by traditional manual feature extraction methods.

\item[3)]

The multimodal feature integration and ACS prediction. In this part, we leverage attention mechanisms to integrate embedded tabular data features with extracted time-series features from the Patched RWKV module. By generating an attention map, the model assigns weights to different features, aligning data from various types and providing explanations for the model’s decisions. This integration enhances the model's predictive accuracy and interpretability, ensuring that the combined information from different data sources contributes effectively to the ACS risk prediction. 

\end{enumerate}

\subsection{Input Embedding}

In this work, the input embedding consists of two parts: embedding clinical tabular data and embedding air pollution time series data. The specific dimensions and processes are detailed in Section 4.4.2 (Implementation Details).

Embedding clinical tabular data involves several preprocessing steps. First, missing values (approximately 2\% in our dataset) were handled using K-Nearest Neighbours (KNN) imputation \cite{pujianto2019k} (Details regarding the selection of KNN and the sensitivity analysis are provided in Section 4.4.2). Next, categorical variables were converted into a numerical format using One-Hot encoding. Subsequently, all numerical features were standardised to have a mean of 0 and a standard deviation of 1 to ensure consistency in scale. Finally, the fully processed feature vector for each patient is embedded into a dense vector representation $X_{tab} \in \mathbb{R}^{D}$ using a multi-layer perceptron (MLP). This embedded vector serves as the input for the multimodal integration stage.

Embedding air pollution time series data involves segmenting the normalised time series into patches and then projecting these patches into an embedding space, resulting in tokens $X_{ts} \in \mathbb{R}^{N \times T \times C}$. This process is described in detail in the following sections (3.3.1 and 3.3.2).

\subsection{PatchRWKV for Time-Series Feature Extraction}

The PatchRWKV module is a core component of the TabulaTime model, designed to handle and extract features from multivariate time-series data, such as air pollution data. This module allows the model to manage both short-term spikes and long-term trends, providing a comprehensive representation of temporal dependencies crucial for accurate analysis. It integrates Recurrent Neural Networks (RNNs) and attention mechanisms to efficiently process long sequence data while maintaining linear computational complexity. The PatchRWKV consists of three key components: Patching, Patch Embedding, and the RWKV encoder. The rationale behind the model design is as follows:

\begin{enumerate}
\item [1)]
Patching: Patching segments time-series data into fixed-size patches, allowing the model to focus on local temporal patterns within each patch and effectively capture short-term dependencies and variations. This enhances locality and semantic understanding, helping the model efficiently learn from short-term dependencies.

\item [2)]
Patch Embedding: The patch embedding component converts the patched data into a format that the model can process. This step ensures that the data is prepared for deeper analysis by the subsequent components.

\item [3)]

RWKV Encoder: Multivariate time series data consists of multiple channels, each representing different types of measurements taken over time. To extract effective feature representations, it is crucial to understand the relationships between different time periods (Time-Mixing) and between different features (Channel-Mixing). The RWKV encoder serves as the backbone architecture of the model, incorporating Time-Mixing and Channel-Mixing to enhance the model’s ability to detect significant patterns and improve predictive performance. Time-Mixing involves using a linear combination of current and previous time steps along with a multi-head Receptance Weighted Key Value (RWKV) operator to capture temporal dependencies across different time patches. This ensures that the model effectively learns the relationships between data points over various time steps, which is crucial for understanding how past events influence future outcomes in a time series. Channel-Mixing is the process of using a vector of all the features from the time series and projecting it into the embedding space. This approach combines information across different channels using robust non-linear operations. Channel-Mixing helps the model learn the relationships between different types of data collected at the same time step, enhancing its ability to interpret multivariate time series data effectively.
\end{enumerate}

The following sections detail these components.

\subsubsection{Patching}


The patching process consists of two stages: normalisation and patching. Instance normalisation is applied first to each univariate time-series pollution dataset to mitigate distribution shift effects. Each normalised time series is then tokenised into patches. In this study, using hourly data, we set the patch size P and stride S based on capturing daily patterns (details in Section 4.4.2). These patches are treated as input tokens by the subsequent RWKV encoder.


\subsubsection{Patch Embedding}

Following normalisation and patching, each patch (a vector of dimension P) is linearly projected into a higher-dimensional embedding space (dimension C). These embedded patch representations form the time-series input tokens $X_{ts}$ for the PatchRWKV module, structured as $X_{ts} \in \mathbb{R}^{N \times T \times C}$, where N is the number of pollution indicators, T is the number of patches (days), and C is the latent feature dimension. The specific dimensions used are detailed in Section 4.3.2.

\subsubsection{RWKV (Receptance Weighted Key Value) Encoder}

\begin{figure}[!htbp]
\centering
\includegraphics[width=0.9\textwidth]{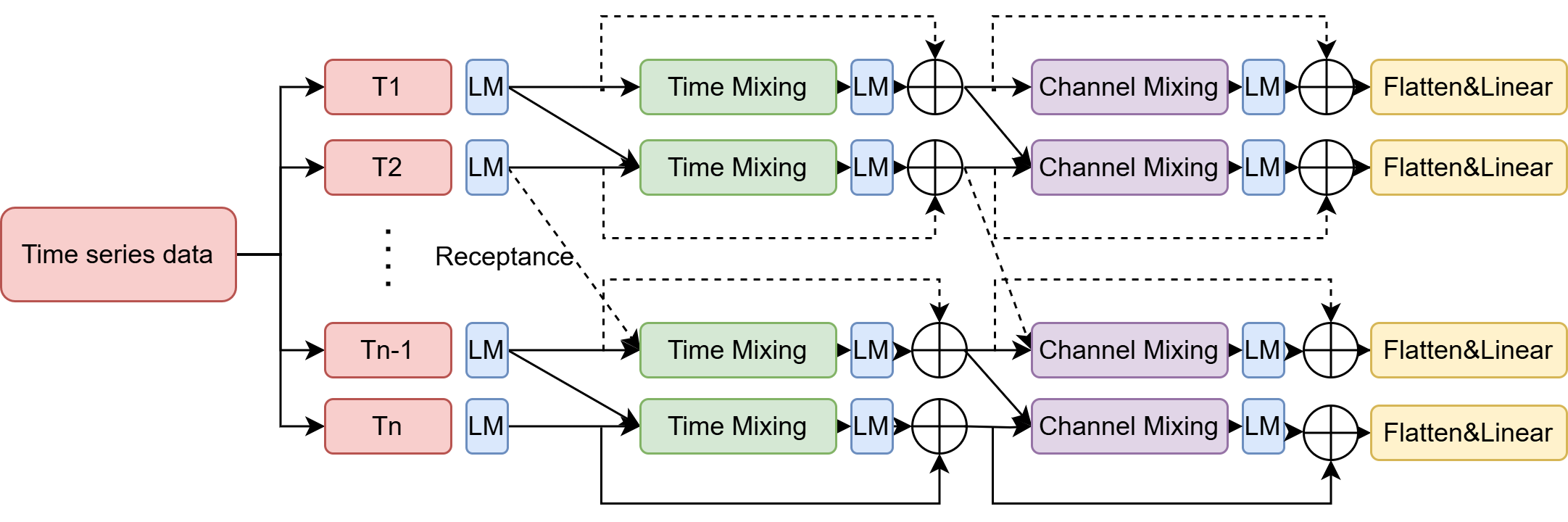}
\caption{Architecture of the RWKV Encoder.}
\label{fig:2}
\end{figure}

\begin{figure}[!htbp]
\centering
\includegraphics[width=0.9\textwidth]{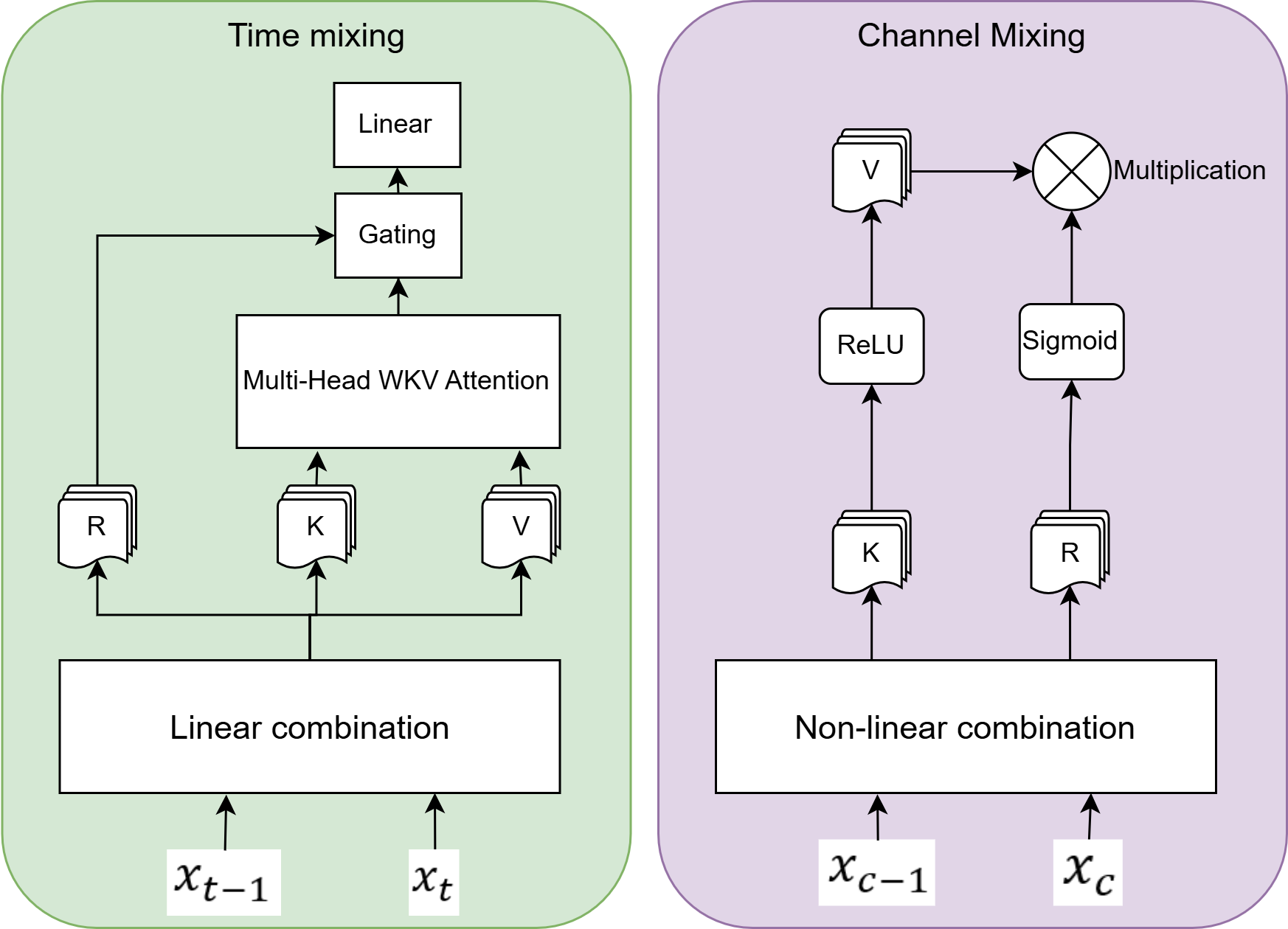}
\caption{Architecture of the time and channel mixing.}
\label{fig:3}
\end{figure}

The RWKV architecture is designed to learn feature representations of time-series data. It consists of stacked residual blocks, each containing a Time-Mixing and a Channel-Mixing sub-block ({Figure~\ref{fig:2}} and {Figure~\ref{fig:3}}). These sub-blocks use recurrent structures to incorporate past information effectively.

\paragraph{Time Mixing Module}

To achieve time mixing in RWKV, the model interpolates between the inputs of the current and previous time-steps. Given an input feature of \(x_{t}\) and previous step \(x_{t - 1}\), a linear projection of the combination of the shifted previous step and the current step is performed using the projection matrix within the block. This process involves creating a weighted blend of~\(x_{t}\)\hspace{0pt}~and~\(x_{t - 1}\)\hspace{0pt}~to capture temporal dependencies effectively.

\begin{equation}
\begin{aligned}
& R_t=W_r \cdot\left(\mu_r \odot x_t+\left(1-\mu_r\right) \odot x_{t-1}\right) \\
& K_t=W_k \cdot\left(\mu_k \odot x_t+\left(1-\mu_k\right) \odot x_{t-1}\right) \\
& V_t=W_v \cdot\left(\mu_v \odot x_t+\left(1-\mu_v\right) \odot x_{t-1}\right)
\end{aligned}
\end{equation}

Where \emph{W} means the weight signifying the positional weight decay vector, a trainable parameter within the model. \emph{\textbf{R}} is the receptance vector acts as the receiver of past information. \emph{\textbf{K}} is the Key vector performs a role analogous to K in traditional attention mechanisms. \emph{\textbf{V:}} The Value vector functions similarly to V in conventional attention processes.

Then a Multi-head WKV Operator is used to operate the attention mechanism but with a linear time and space complexity. This recurrent behaviour in RWKV is articulated through the time-dependent update of the WKV vectors. The formula of single head WKV operator is given by.

\begin{equation}
w k v_t=\operatorname{diag}(u) \cdot k_t^T \cdot v_t+\sum_{i=1}^{t-1} \operatorname{diag}(w)^{(t-1-i)} \cdot k_i^T \cdot v_i
\end{equation}

where \emph{\textbf{w}} and \emph{\textbf{u}} are two trainable parameters. The parameter \emph{u} is a bonus that rewards the model for encountering a token for the first time, specifically the current token. This ensures the model pays more attention to the current token, preventing any potential degradation of \emph{\textbf{w}}. Another important parameter is \emph{\textbf{w}}, which is a channel-wise time decay vector per head. Furthermore, we transform parameter \emph{w} within the range (0,1), ensuring that \(diag(w)\) represents a contraction matrix.

Like the transformer structure, we use the multi-head WKV to enhance the model's capacity which is formally described by the following equation:

\begin{equation}
M H_{w k v_t}=\operatorname{Concat}\left(w k v_t^1, \ldots, w k v_t^h\right)
\end{equation}

Where h is the number of heads. Finally, an output gating mechanism controls the flow of information from the recurrent unit to the next layer or the final output. The output gating is implemented using the sigmoid ($\delta$) of $r_t$. The output vector \(o_{t}\) per head is given by:

\begin{equation}
o_t=\left(\delta(r_t) \cdot w k v_t\right) W_o
\end{equation}

where $W_o$ represents the weight matrix for the output projection following the gating and attention calculation.

\paragraph{Channel Mixing Module}

In the channel-mixing block, channels are mixed by strong non-linear operations as follow:

\begin{equation}
\begin{gathered}
k_t^{\prime}=W_g^{\prime} \cdot\left(\mu_k^{\prime} \odot x_t+\left(1-\mu_k^{\prime}\right) \odot x_{t-1}\right) \\
r_t^{\prime}=W_r^{\prime} \cdot\left(\mu_r^{\prime} \odot x_t+\left(1-\mu_r^{\prime}\right) \odot x_{t-1}\right) \\
v_t^{\prime}=\operatorname{ReLU}^2\left(k_t^{\prime}\right) \cdot W_v^{\prime} \\
o_t^{\prime}=\operatorname{\delta}\left(r_t^{\prime}\right) \odot v_t^{\prime}
\end{gathered}
\end{equation}

where we adopt the squared ReLU activation function to enhance the non-linearity.

\subsection{Multimodal Feature Integration and Prediction}

\begin{figure}[!htbp]
\centering
\includegraphics[width=0.9\textwidth]{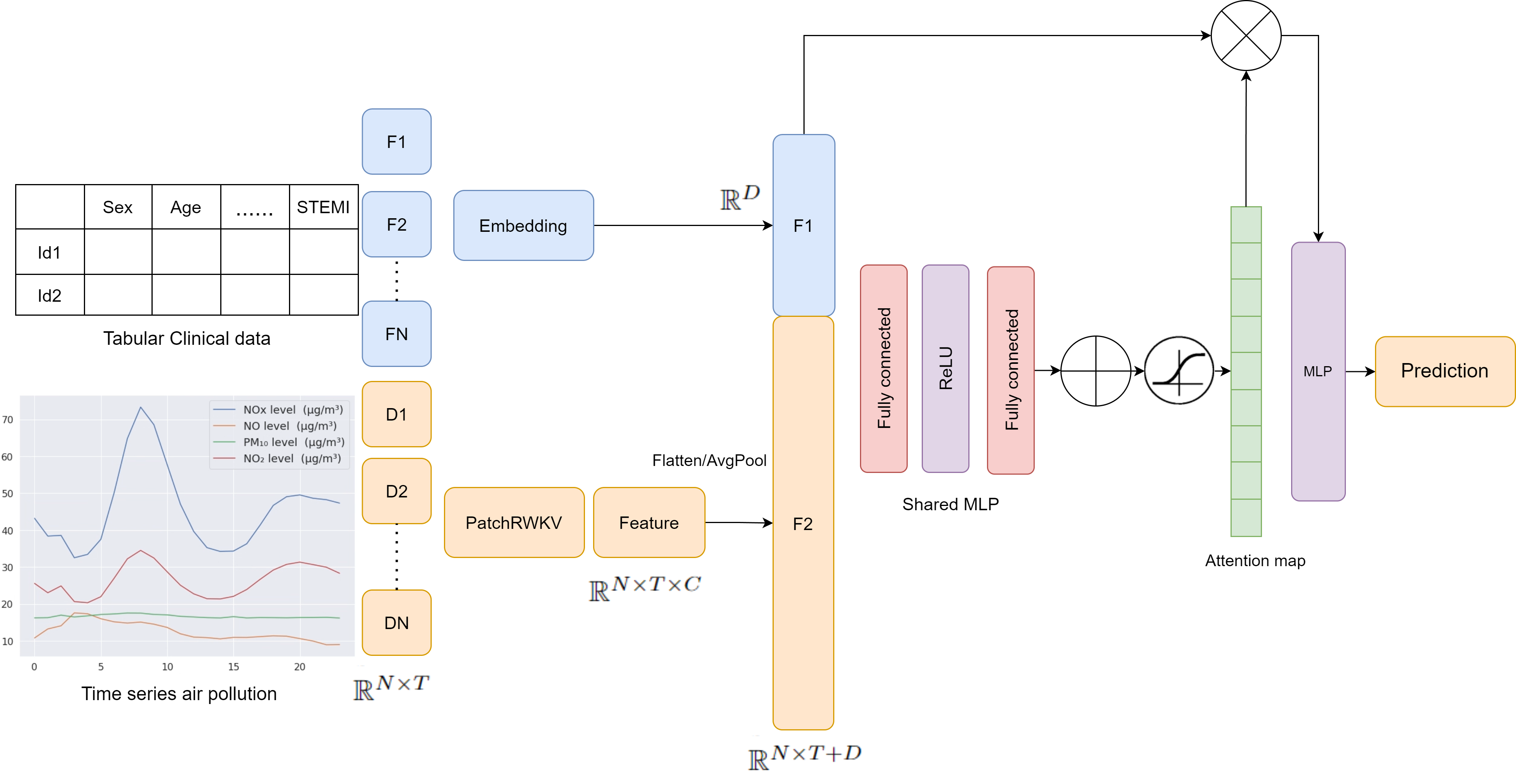}
\caption{An illustrative example of using attention mechanisms for
multimodal feature integration.}
\label{fig:4}
\end{figure}

Feature integration and prediction in the TabulaTime model combine the embedded clinical tabular data vector with the sequence of extracted features from the time-series air pollution data for ACS risk prediction({Figure~\ref{fig:4}}). This model integrates tabular data embedding feature representations of time-series data into an attention module, enabling the network to learn attention maps conditionally based on the tabular data. This integration enhances the network's ability to pinpoint what, where, and when to focus on in the data, leading to improved performance in tasks involving both time series and tabular data.

Let $X_{tab} \in \mathbb{R}^{D}$ be the embedded vector representation of the clinical tabular data, processed as described in Section 3.2 and projected to the dimension D. $X_{ts} \in \mathbb{R}^{N \times T \times C}$ represent the sequence of feature vectors output by the PatchRWKV module (where N is the number of air pollution types, T is the effective sequence length (number of days), and C is the feature dimension). This tensor is first flattened across the N and T dimensions, and then global average pooling is applied across the flattened dimension, resulting in a 1D vector $X_{ts\_pooled} \in \mathbb{R}^{N \times T}$.



The model integrates these representations using an attention mechanism. Give the input \(X \in \mathbb{R}^{D+N\times T}\), the feature integration can be formulated as:

\begin{equation}
\hat{X}=\sigma\left(W_2 \cdot \delta\left(W_1 \cdot X\right)\right) \cdot X
\end{equation}

Where $W_1$ and $W_2$  are the weight matrices of the two fully connected layers. $\delta$ denotes the ReLU activation function.$\sigma$ denotes the sigmoid activation function. $\sigma\left(W_2 \cdot \delta\left(W_1 \cdot X_c\right)\right)$ refers to the generated attention map. As a final step, a MLP is added at the end of the model to predict the ACS risk.

This process ensures that the model effectively integrates and leverages both the tabular and time-series data, leading to more accurate and reliable ACS risk predictions.

\section{Experimental Evaluation}

\subsection{Experimental Design}

To evaluate the efficacy of the TabulaTime model, we conduct three comprehensive experiments, each designed to assess distinct aspects of its performance:

\begin{enumerate}

\item [1)] Model Performance Analysis

This experiment evaluates the predictive accuracy of TabulaTime for Acute Coronary Syndrome (ACS) risk. We test the model using varying air pollution and climate data durations (3, 7, 10, and 14 days) to identify the optimal time window. Subsequently, we compare its performance against traditional machine learning models, including Random Forest, LightGBM \cite{ke2017lightgbm}, and CatBoost\cite{prokhorenkova2018catboost}, under two scenarios: (a) incorporating environmental data, and (b) excluding environmental data. These comparisons quantify the significance of environmental factors in enhancing predictive accuracy. Additionally, the generalisability of TabulaTime are assessed across varying time periods and air conditions.

\item[2)]  PatchRWKV Module Evaluation

This experiment focus on the PatchRWKV module's capability in feature extraction from time-series data. Using three publicly available datasets, we benchmark its performance in both classification and forecasting tasks, highlighting its ability to efficiently capture complex temporal patterns. We select several State-of-the-art models and referenced their results, which includes most recent and extensive empirical studies on time-series. The selected models include:

\textbf{Multi-Layer Perceptrons (MLPs) -based models}: LightTS \cite{Zhang2022Less}, LightTS is a MLPs model which is straightforward and computationally efficient, suitable for real-time and large-scale applications. 

\textbf{Convolutional Neural Networks (CNNs)-based models}: TimesNet \cite{Wu2023TimesNet}, TimesNet is a CNN model which is efficient in capturing local patterns and hierarchical features in time-series data, providing a solid comparison for the feature extraction capabilities of the PatchRWKV module .

\textbf{Transformers-based models}: Autoformer\cite{Wu2022Autoformer},  PatchTST \cite{Nie2023Time}. Transformer have revolutionised sequence modelling with their self-attention mechanisms, excelling in capturing long-range dependencies. Both Autoformer and PatchTST represent state-of-the-art approaches in time-series analysis with their decomposition mechanism and patching mechanism.

\textbf{Recurrent Neural Networks (RNNs) -based models}: Gated Recurrent Unit (GRU) \cite{Cho2014Learning} and Long Short-Term Memory (LSTM) \cite{Hochreiter1997Long} are designed for sequence data and known for modelling temporal dependencies effectively. These models provide a comparison for the recurrent structures within the PatchRWKV module.

    \item[3)]  Feature Importance and Model Interpretability

This study aims to evaluate the predictive performance of a novel deep learning framework (TabulaTime) that integrates selected clinical risk factors  and time-series environmental data for differentiating ACS presentation (STEMI vs. NSTEMI). In this experiment, we assess the model's ability to identify significant features and their contributions to the prediction of ACS risk. By analysing feature importance scores, we identified the key predictors and their relative contributions, enhancing the model’s interpretability and offering valuable insights for targeted interventions. The feature importance is calculated using permutation feature importance\cite{Fisher2019All}. This method assesses the increase in the model’s prediction error when a feature’s values are shuffled. If shuffling a feature increases the model error, the feature is considered important, as the model relies on it for predictions. Conversely, if shuffling does not affect the model error, the feature is deemed unimportant, as the model ignores it for predictions. We first calculate feature importance as the drop in the model’s validation metric when a feature value is randomly shuffled. Then, we calculate step importance to show the significance of each time period. 
    
\end{enumerate}

\subsection{Dataset Description}

The study utilises three primary datasets to evaluate the performance of the TabulaTime model: 
1) Salford MINAP dataset, 2) Air Pollution and Climate Dataset and 3) Additionally, to evaluate  the performance of PatchRWKV, we have also  used three public available time series datasets including Weather \cite{Bahdanau2016Neural}, Heartbeat Monitoring Dataset and Self-Regulation of Slow Cortical Potentials (SCP) Dataset.

\subsubsection{Salford MINAP Dataset}

The Myocardial Ischaemia National Audit Project (MINAP) is a domain within the National Cardiac Audit Programme (NCAP) that contains information about the care provided to patients who are admitted to hospital with acute coronary syndromes (heart attack). It is collected and analysed to illustrate the ‘patient journey’ from a call to the emergency services or their self-presentation at an Emergency Department, through diagnosis and treatment at the hospital, to the prescription of preventive medications on discharge. A pseudonymised dataset was obtained under ethical approval (REC reference: 22/YH/0250) after confidentiality advisory group review (CAG reference: 22/CAG/0155) for use in this study.

In this work, 21 risk factors including patient demographics, clinical history, medications, vital signs and diagnostic indicators are selected to predict the ACS risk and differentiate between STEMI and NSTEMI based on their relevance and significance in clinical and environmental contexts. These 21 selected risk factors represent a combination of data types pertinent to ACS risk and presentation:

\textbf{Baseline and Historical Data:} Including patient demographics (age, sex, BMI derived from height and weight), documented pre-existing conditions and medical history (e.g., Previous AMI, Previous Angina, Hypertension, Hypercholesterolaemia, Diabetes, Chronic Renal Failure, Heart Failure, Cerebrovascular Disease, Peripheral Vascular Disease, Asthma/COPD, Family History of CHD), prior interventions (Previous PCI, Previous CABG), relevant lifestyle factors (Smoking Status), and chronic medication usage (ACEI or ARB, Statin). These factors reflect the patient's condition prior to the specific ACS event which are crucial as they provide a comprehensive view of a patient's medical history and risk factors associated with cardiovascular diseases\cite{Rhyou2018Clinical,Yunyun2014Analysis}.




\textbf{Admission Data:} Including key physiological measurements typically obtained upon presentation or during the initial assessment for the ACS event, such as Systolic Blood Pressure and serum Creatinine levels. Additionally, the time from symptom onset to therapy was calculated by determining the duration between the recorded 'Onset of Symptoms' time and the 'Admission Date/Time' for each patient.

This comprehensive variable set allows the model to learn from both long-term antecedent risk factors and indicators present at the time of the acute clinical presentation, in conjunction with the preceding environmental data.The detailed risk factors are shown in Table~\ref{tab:1}.

Based upon electrical heart tracings – electrocardiograms or ECGs recorded during a heart attack, patients are diagnosed as having suffered either ST-elevation myocardial infarction (STEMI) or non-ST elevation myocardial infarction (NSTEMI).

\begin{table*}[!htbp]
    \begin{adjustbox}{max width=\textwidth}
    \begin{tabular}{ll}
    \hline
    Medical Conditions and Measurements                                                        & Counting and calculation formulae   \\ \hline
    Previous AMI (Acute Myocardial Infarction)                                                 & Yes 381/No 597                      \\
    Previous Angina                                                                            & Yes 470/No 511                      \\
    Hypertension                                                                               & Yes 591/No 389                      \\
    Hypercholesterolaemia                                                                      & Yes 623/No 358                      \\
    Peripheral Vascular Disease                                                                & Yes 103/No 877                      \\
    Cerebrovascular Disease                                                                    & Yes 125/No 855                      \\
    Asthma/COPD                                                                                & Yes 274/No 704                      \\
    Chronic Renal Failure                                                                      & Yes 143/No 837                      \\
    Heart Failure                                                                              & Yes 187/No 792                      \\
    Smoking Status                                                                             & Ex 384/Current smoker 251/Never 343 \\
    Diabetes                                                                                   & Yes 298/No 683                      \\
    Previous PCI (Percutaneous Coronary Intervention)                                          & Yes 184/No 797                      \\
    Previous CABG (Coronary Artery Bypass Graft)                                               & Yes 99/No 880                       \\
    ACEI or ARB (Angiotensin-Converting Enzyme Inhibitors or Angiotensin II Receptor Blockers) & Yes 471/No 507                      \\
    Systolic BP (Blood Pressure)                                                               & 137.5 ± 27.1                        \\
    Height                                                                                     & 166.6 ± 10.3                        \\
    Weight                                                                                     & 78.1 ± 20.57                        \\
    BMI                                                                                        & $\mathrm{BMI}=\frac{\text { Weight }}{(\text { Height })^2}$                                    \\
    Family History of CHD (coronary heart disease)                                             & Yes 176 /No 785                     \\
    Creatinine                                                                                 & 112.4 ± 90.3                        \\
    Statin                                                                                     & Yes 592/No 386                      \\
    Age                                                                                        & 71.1 ± 13.7                         \\
    SymptomToAdmissionTime                                                                     & -8673 to 15131 Hours           \\
    STEMI                                                                                      & 123/587                             \\ \hline
    \end{tabular}
    \end{adjustbox}
    \caption{Features used in MINAP dataset.}
    \label{tab:1}
    \end{table*}

\subsubsection{Air Pollution and Climate Dataset}

This air pollution dataset contains hourly measurements of various air pollutants collected from the Salford Eccles monitoring station (UKA00339) in the United Kingdom. The time range covered by the data is from January 1, 2016, to December 31, 2019. The primary indicators monitored include levels of Nitrogen Oxides (NO$_x$), Nitric Oxide (NO), Particulate Matter (PM$_{10}$), and Nitrogen Dioxide (NO$_{2}$), all measured in micrograms per cubic meter (µg/m³).

In addition to air pollution, hourly climate data, specifically temperature (measured in degrees Celsius), was incorporated into the analysis. This data was obtained from the Meteostat (https://meteostat.net/en/) public data source, using readings from a weather station geographically close to the study area (Manchester Airport station, identifier 03330). The climate data covers the same temporal range as the air pollution datasets to ensure alignment for the multimodal analysis. Temperature data was included as it is a known environmental factor that can influence cardiovascular events.

Crucially, both the air pollution and climate datasets were temporally aligned with the clinical data from the Salford MINAP dataset. For each patient's admission date, the corresponding preceding environmental data (e.g., for the 10 days prior) was extracted and matched, ensuring that the model analyses the environmental conditions leading up to the ACS event.
% ...

\subsubsection{Public Datasets}

Weather\cite{Bahdanau2016Neural}, which is recorded every 10 minutes through the whole of the year 2020,  contains 21 meteorological indicators, such as air temperature, humidity, etc. This dataset is used for time series forecasting.
Heartbeat Monitoring Dataset\cite{Kanani2020ECG} is used to classify and monitor heartbeats. The dataset consists of sequences of heartbeat signals, often represented as electrocardiogram (ECG) readings, which need to be classified into different categories such as normal or abnormal heartbeats.
Self-Regulation of Slow Cortical Potentials (SCP) Dataset \cite{Birbaumer1999spelling} is used for studying self-regulation of brain activity, specifically the ability to control slow cortical potentials (SCPs), which are slow voltage changes in the brain's electrical activity. The dataset includes sequences of brain activity measurements, which need to be classified to understand patterns related to self-regulation capabilities.

\subsection{Performance Metrics}

In this study, we evaluate the performance of our proposed TabulaTime
model using several key metrics that are standard in the fields of
classification and forecasting. These metrics provide a comprehensive
view of the model\textquotesingle s predictive capabilities and its
effectiveness in handling both classification tasks (e.g., predicting
STEMI or NSTEMI) and forecasting tasks (e.g., forecasting air pollution
levels).

\subsubsection{Classification Metrics}

For classification tasks such as distinguishing between STEMI and NSTEMI cases, the following metrics were used:

\textbf{Accuracy}: Accuracy is a widely used metric that represents the proportion of correctly predicted instances among the total instances.

\begin{equation}
\text { Accuracy }=\frac{\mathrm{TP}+\mathrm{TN}}{\mathrm{TP}+\mathrm{FP}+\mathrm{FN}+\mathrm{TN}}
\end{equation}

\textbf{Precision:} Proportion of true positive predictions out of all positive predictions made by the model, highlighting prediction reliability.

\textbf{Recall:} Proportion of true positive predictions out of all actual positive cases, indicating the model's ability to detect positive instances.

\begin{equation}
\text { Precision }=\frac{\mathrm{TP}}{\mathrm{TP}+\mathrm{FP}}
\end{equation}

\textbf{F1-Score:} Harmonic mean of Precision and Recall, balancing both metrics for a comprehensive assessment of classification performance.

\begin{equation}
\text { Recall }=\frac{\mathrm{TP}}{\mathrm{TP}+\mathrm{FN}}
\end{equation}

\textbf{ROC Curve and AUC (Area Under the Curve):} The ROC curve is a graphical representation of a classification model's performance across all classification thresholds. It plots the True Positive Rate (TPR) against the False Positive Rate (FPR). The AUC provides a single scalar value to summarise the model's performance; a higher AUC indicates better model performance.

\subsubsection{Forecasting Metrics}

\textbf{Mean Squared Error (MSE) and Mean Absolute Error (MAE)}: MSE measures the average squared difference between the actual and predicted values. It is particularly sensitive to large errors, making it a suitable metric for assessing the accuracy of continuous predictions. MAE is another metric used to measure the average magnitude of errors in a set of predictions, without considering their direction (positive or negative). It provides a clear interpretation of the average error in the same units as the data. MSE and MAE are calculated as:

\begin{equation}
\mathrm{MSE}=\frac{1}{n} \sum_{i=1}^n\left(\hat{y}_i-y_i\right)^2
\end{equation}

\begin{equation}
\mathrm{MAE}=\frac{1}{n} \sum_{i=1}^n\left|\hat{y}_i-y_i\right|
\end{equation}

Where $\hat{y}_i$ is the predicted value and $y_i$ is the actual value.

\subsection{Experimental configuration}

\subsubsection{Data Preprocessing}

Data preprocessing is a critical step before model training. For the clinical tabular data from the MINAP dataset, several steps were taken. Categorical variables were converted into numerical format using One-Hot encoding. Numerical features were standardised to have a mean of 0 and a standard deviation of 1 to ensure consistency in data scale and improve model convergence.

Handling missing values is essential for robust model performance. In our dataset, missing values were relatively sparse, constituting approximately 2\% of the clinical data. Our primary analysis employed K-Nearest Neighbours (KNN) imputation. This method estimates missing values by referencing the five most similar complete records ('neighbour') based on other available features. KNN imputation is advantageous as it preserves local data structure and inter-feature relationships, making it particularly suitable for datasets with a low proportion of missing data like ours.

To justify the selection of KNN, we conducted a sensitivity analysis comparing three common imputation strategies:
\begin{enumerate}
    \item Mean imputation: Replacing missing values with the mean of the respective feature.
    \item K-Nearest Neighbours (KNN) imputation (k=5): The primary method used in this study.
    \item Multivariate Imputation by Chained Equations (MICE): An iterative regression-based method that models each variable with missing values as a function of the other variables.
\end{enumerate}

The comparative performance metrics (Accuracy and AUC) for the TabulaTime model using data imputed by these three methods are summarised in Table~\ref{tab:imputation_comparison}.

\begin{table}[!htbp]
\begin{adjustbox}{max width=\textwidth}
\centering
\begin{tabular}{llllll}
\hline
\textbf{Imputation Method} & \textbf{Accuracy} & \textbf{AUC} \\ \hline
Mean Imputation            & 0.79              & 0.87         \\
MICE                       & 0.81              & 0.89         \\
KNN Imputation (k=5)       & \textbf{0.82}     & \textbf{0.91} \\ \hline
\end{tabular}
\end{adjustbox}
\caption{A sensitivity analysis comparing three primary imputation methods was conducted to support the choice of KNN.}
\label{tab:imputation_comparison}
\end{table}

These results indicate that KNN imputation achieved the best overall model performance in terms of both Accuracy and AUC, with MICE yielding comparable results. Both KNN and MICE significantly outperformed simple mean imputation. This sensitivity analysis confirms that the model’s performance is robust across different sophisticated imputation techniques and supports our decision to adopt KNN imputation as the primary method for handling missing data in this study.

\subsubsection{Implementation Details}

All experiments were implemented using Python 3.9. The deep learning models, including TabulaTime and its PatchRWKV module, were built using the PyTorch (version 2.5). Traditional machine learning models (Random Forest, LightGBM, CatBoost) were implemented using the Scikit-learn library. The source code for the TabulaTime framework and the experiments conducted in this study is publicly available at: \url{https://gitlab.com/han-research/tabulatime}.

\textbf{Embedding Dimensions:}
The specific dimensions used for the input embeddings are as follows:
\begin{itemize}
    \item \textbf{Tabular embedding ($X_{tab}$):} The pre-processed clinical tabular data (including 21 clinical features and 10 air pollution statistical features) is embedded into a dense vector of dimension D = 31 using an MLP.
    \item \textbf{Patch embedding ($X_{ts}$):} Hourly environmental data is segmented into non-overlapping daily patches (Patch size P = 24, Stride S = 24). Each patch is flattened and linearly projected into an embedding space with dimension C = 128. The resulting tensor has shape $X_{ts}\in\mathbb{R}^{N\times T\times C}$, where N is the number of environmental indicators (channels) and T is the number of patches (days, e.g., 10 for a 10-day sequence).
\end{itemize}

\textbf{Training Configuration:}

For model training and evaluation, the data was partitioned into training (80\%) and testing (20\%) subsets, ensuring temporal order was maintained where applicable. A validation subset (10\% of the training data) was used for hyperparameter tuning and early stopping. The TabulaTime model, including the PatchRWKV module (configured with 4 layers and an embedding dimension C=128), was trained using the Adam optimiser with a learning rate of 0.001 and a batch size of 16. The training process was run for a maximum of 100 epochs, employing early stopping with a patience of 10 epochs based on the validation loss (Cross-Entropy Loss for classification tasks) to prevent overfitting. Baseline machine learning models (RF, LightGBM, CatBoost) were trained using their default hyperparameters as provided by Scikit-learn, representing standard configurations.


\section{Results}

\subsection{The Performance Evaluation of TabulaTime in ACS Prediction}

The evaluation of the TabulaTime model was conducted through a series of experiments to assess its predictive performance for description of Acute Coronary Syndrome (ACS) risk. The experiments were designed to evaluate the influence of time-series air pollution data, compare TabulaTime with existing machine learning models, and analyse its generalisation capabilities under varying conditions.

\subsubsection{Impact of Time-Series Length for ACS Prediction}

The model was tested with input sequences corresponding to varying preceding durations (3, 7, 10, and 14 days). For the optimal 10-day window, this involved processing 10 consecutive 24-hour patches derived from the hourly air pollution data. As shown in Table~\ref{tab:2}, the 10-day time series achieved the best results, with an accuracy of 82.3\% and an AUC of 0.91, outperforming other time lengths across all metrics. This suggests that a 10-day window strikes a balance between capturing essential trends and avoiding noise introduced by excessive data. Shorter windows (3 and 7 days) lacked critical information, while longer windows (14 days) diluted key patterns.

\begin{table*}[!htbp]
\begin{adjustbox}{max width=\textwidth}
\begin{tabular}{llllll}
\hline
\multicolumn{1}{c}{\textbf{Time   Series Length (Days)}} & \multicolumn{1}{c}{\textbf{Accuracy}} & \multicolumn{1}{c}{\textbf{Precision}} & \multicolumn{1}{c}{\textbf{Recall}} & \multicolumn{1}{c}{\textbf{F1-Score}} & \multicolumn{1}{c}{\textbf{AUC}} \\ \hline
3 Days                                                   & 0.815                                 & 0.809                                  & 0.822                               & 0.815                                 & 0.882                            \\
7 Days                                                   & 0.820                                 & 0.817                                  & 0.825                               & 0.821                                 & 0.894                            \\
10 Days                                                  & \textbf{0.823}                        & \textbf{0.820}                         & \textbf{0.830}                      & \textbf{0.825}                        & \textbf{0.914}                   \\
14 Days                                                  & 0.819                                 & 0.813                                  & 0.820                               & 0.816                                 & 0.894                            \\ \hline
\end{tabular}
\end{adjustbox}
\caption{Model prediction performance with time-series air pollution data of varying lengths.}
\label{tab:2}
\end{table*}

\subsubsection{Comparison With Existing Machine Learning Models for ACS Risk Prediction}

In this section, we evaluated the performance of the proposed method for ACS risk prediction, compared it with several state-of-the-art machine learning models, including Random Forests (RF), Light Gradient Boosting Machines(LightGBM), and CatBoost. The aim of this task was to predict whether a patient had STEMI or NSTEMI under different scenarios with and without air pollution data included. The accuracies of the proposed method and the existing models were reported in Table~\ref{tab:3}.

Random Forests (RF) is an ensemble learning technique that constructs multiple decision trees and outputs from either the mode of the classes (for classification) or the mean prediction (for regression) from the individual trees. This method is robust against overfitting and handles a large number of input features effectively. Both LightGBM and CatBoost are gradient boosting algorithms that are particularly well-suited for structured/tabular data, offering high performance and efficiency.

\begin{table*}[!htbp]
\begin{adjustbox}{max width=\textwidth}
\begin{tabular}{lllll}
\hline
Methods              & Random Forest (RF) & LightGBM & Catboost & The proposed TabulaTime \\ \hline
W/O air pollution    & 0.6                & 0.645    & 0.665    & 0.753                   \\
With air   pollution & 0.627              & 0.655    & 0.688    & 0.829                   \\ \hline
\end{tabular}
\end{adjustbox}
\caption{Comparison of classification accuracy (\%) for STEMI vs. NSTEMI prediction using traditional machine learning models (Random Forest, LightGBM, CatBoost) and the proposed TabulaTime method, under scenarios with and without air pollution data.}
\label{tab:3}
\end{table*}

The performance metrics are provided for two scenarios: without considering air pollution and with considering air pollution. In both scenarios, TabulaTime consistently outperforms RF, LightGBM and Catboost. When air pollution data was included, TabulaTime significantly outperformed the other models. It performed 32.2\% better than Random Forest, 27.5\% better than LightGBM, and 20.5\% better than CatBoost. Without the inclusion of air pollution data, TabulaTime performs 25.5\% better than Random Forest, 16.7\% better than LightGBM, and 13.2\% better than CatBoost.

\begin{figure}[!htbp]
\centering
\includegraphics[width=0.9\textwidth]{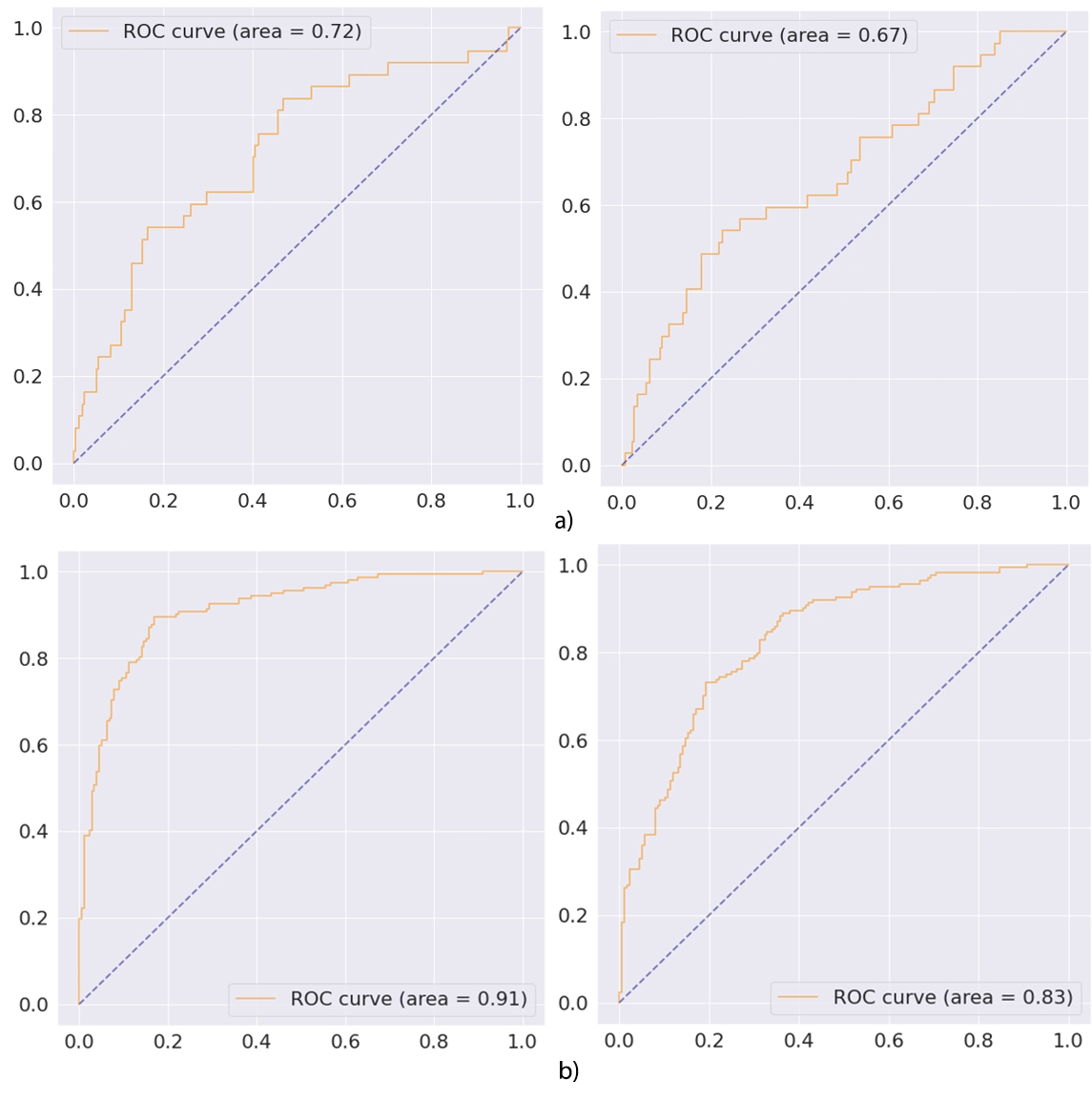}
\caption{ROC curve of Catboost(a) and TabulaTime(b) with and without air pollution.}
\label{fig:5}
\end{figure}

Both models showed an improvement when air pollution is included as a factor. The performance of RF improves by 4.5\% (from 0.6 to 0.627), LightGBM improved by 1.6\% (from 0.645 to 0.655) and Catboost improves by 3.5\% \% (from 0.665 to 0.688). TabulaTime\textquotesingle s performance improved by 10.1\% (from 0.753 to 0.829). This indicated that incorporating air pollution data enhanced the predictive capabilities of both models, but the improvement was more pronounced for TabulaTime. {Figure~\ref{fig:5}} reported the Area Under the Receiver Operating Characteristic Curve (AUC-ROC) for the best-performing machine learning model (Catboost) and the proposed TabulaTime. The results confirm that the model performance improved with the inclusion of air pollution indicators. The ROC of the Catboost increased from 0.67 to 0.72, while the proposed TabulaTime improved the prediction accuracy from 0.83 to 0.91.

\subsubsection{Generalisation Performance Under Varying Conditions}

The generalisability of a model is a crucial indicator of its performance. Notably, during the COVID-19 lockdown in the UK (March 2020 to December 2021), there was a significant decline in air pollution levels. However, the incidence of heart attacks, particularly NSTEMI, increased, a trend confirmed by various studies \cite{Little2020COVID,Topol2020COVID}.

\begin{figure}[!htbp]
\centering
\includegraphics[width=0.9\textwidth]{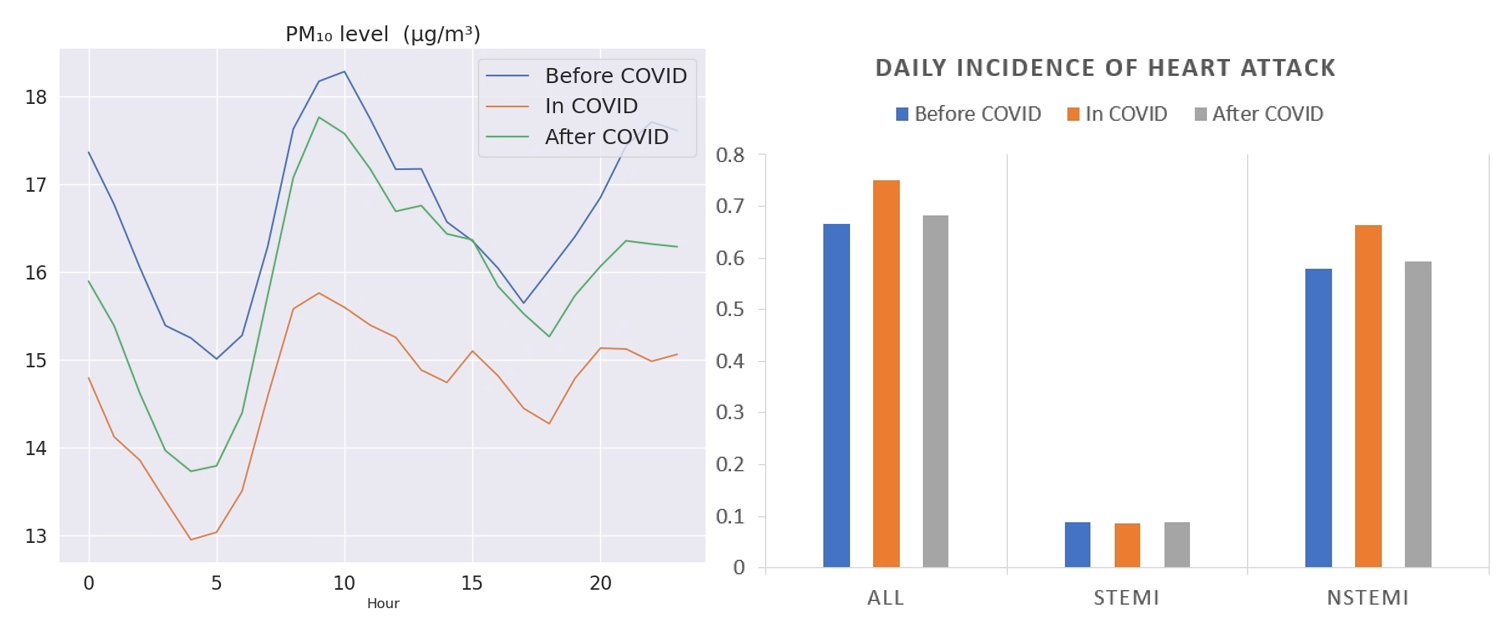}
\caption{Changes in air pollution indicator PM$_{10}$ and heart attack daily
incidence over different periods.}
\label{fig:7}
\end{figure}

We have evaluated the model performance across three distinct periods: pre-COVID (January 2016 to December 2019), during the COVID lockdown (March 2020 to December 2021), and post-COVID lockdown (January 2022 to December 2023). {Figure~\ref{fig:7}} illustrates the PM$_{10}$ levels and the daily incidence of heart attacks during these periods. The model's accuracy for these periods was 0.829, 0.788, and 0.802, respectively, despite significant fluctuations in air pollution levels. This consistency in model accuracy across different conditions indicates that our model remains effective under various scenarios.

\subsection{The Performance of PatchRWKV}

In the second experiment, we evaluated the performance of the proposed time series feature extraction module (PatchRWKV). Two types of tasks were used to evaluate the model's performance: \textbf{Classification and Forecasting}.

\subsubsection{Classification}

The aim of this experiment was to assess the model's ability to learn high-level representations through sequence-level classification in time series data. We selected three multivariate datasets from medical diagnosis including heartbeat monitoring and Self-regulation of Slow Cortical Potentials and ACS prediction.

\begin{table*}[!htbp]
\begin{adjustbox}{max width=\textwidth}
\begin{tabular}{lrrrrrllr}
\hline
Methods            & \multicolumn{1}{l}{LightTS} & \multicolumn{1}{l}{DLinear} & \multicolumn{1}{l}{TimesNet} & \multicolumn{1}{l}{Autoformer} & \multicolumn{1}{l}{PatchTST} & GRU                       & LSTM  & \multicolumn{1}{l}{PatchRWKV} \\ \hline
Heartbeat          & 0.751                       & 0.751                       & 0.756                        & 0.737                          & 0.771                        & 0.722                     & 0.751 & 0.780                         \\
SelfRegulationSCP1 & 0.898                       & 0.873                       & 0.846                        & 0.887                          & 0.904                        & \multicolumn{1}{r}{0.549} & 0.744 & 0.918                         \\
SelfRegulationSCP2 & 0.511                       & 0.505                       & 0.556                        & 0.544                          & 0.567                        & 0.533                     & 0.517 & 0.572                         \\
ACS                & 0.681                       & 0.667                       & 0.676                        & 0.686                          & 0.678                        & 0.676                     & 0.686 & 0.703                         \\ \hline
\end{tabular}
\end{adjustbox}
\caption{Classification accuracy (\%) of various deep learning models on multivariate medical time series datasets including heartbeat monitoring, self-regulation of slow cortical potentials (SCP), and acute coronary syndrome (ACS) prediction.}
\label{tab:4}
\end{table*}

Table~\ref{tab:4} presents the classification accuracy of eight time series processing models across four classification tasks (Heartbeat, SelfRegulationSCP1, SelfRegulationSCP2, ACS). In Heartbeat task, PatchRWKV improved by 0.9\% over PatchTST and showed a significant improvement of 5.8\% over GRU. In SelfRegulationSCP1 and SelfRegulationSCP2 tasks, PatchRWKV (91.80 and 57.20) outperformed the next best method, PatchTST (90.40 and 56.70), by 1.40\% and 0.50 \%. In the ACS classification task, PatchRWKV (70.28) outperformed the LSTM and Autoformer (both 68.58), by 1.7\%.

PatchRWKV consistently outperformed the other models, achieving the highest accuracy across all tasks: Heartbeat: 78\%, SelfRegulationSCP1: 91.8\%, SelfRegulationSCP2: 57.2\% and ACS: 70.28\%. The improvements range from 0.50\% to 36.85\%, indicating that PatchRWKV is a robust and effective method for these classification tasks, especially with a substantial advantage in more complex tasks like SelfRegulationSCP1. Traditional models like GRU and LSTM lag significantly behind newer architectures, indicating a shift in effectiveness towards more advanced methods like PatchTST and TimesNet.

\subsubsection{Forecasting}

The aim of this experiment was to evaluate the performance of the proposed and compared models in time series forecasting tasks for weather and air pollution prediction. Two real-world datasets were used: the Weather Dataset and the Salford Air Pollution Dataset. 

For the Weather Dataset, 144 historical time steps (24 hours) were used to forecast the next 48 time steps (8 hours), for two indicators: pressure and temperature. In the Salford Air Pollution Dataset, 240 historical time steps (10 days) were used to predict the next 48 time steps (2 days) for four indicators:  Particulate Matter (PM$_{10}$), Nitrogen Dioxide (NO$_{2}$), Nitrogen Oxides (NO$_x$) and Nitric Oxide (NO).



\begin{table*}[!htbp]
\begin{adjustbox}{max width=\textwidth}
\begin{tabular}{cccccc}
\hline
\multicolumn{2}{c}{\multirow{2}{*}{\textbf{Methods}}}    & \multicolumn{2}{c}{\textbf{Weather}} & \multicolumn{2}{c}{\textbf{Salford   Air Pollution}} \\ 
\multicolumn{2}{c}{}                                     & MSE               & MAE              & MSE                       & MAE                      \\ \hline
\multirow{2}{*}{\textbf{MLP-based}}        & LightTS    & 0.261             & 0.312            & 0.693                     & 0.511                    \\
                                            & DLinear    & 0.249             & 0.3              & 0.695                     & 0.508                    \\ \hline
\textbf{CNN-based}                          & TimesNet   & 0.259             & 0.287            & 0.78                      & 0.504                    \\ \hline
\multirow{2}{*}{\textbf{Transformer-based}} & Autoformer & 0.338             & 0.382            & 0.707                     & 0.49                     \\
                                            & PatchTST   & 0.231             & 0.266            & 0.685                     & 0.481                    \\ \hline
\multirow{3}{*}{\textbf{RNN-based}}         & GRU        & 0.225             & 0.286            & 0.718                     & 0.524                    \\
                                            & LSTM       & 0.199             & 0.262            & 0.722                     & 0.52                     \\
                                            & PatchRWKV  & 0.177             & 0.264            & 0.674                     & 0.473                    \\ \hline
\end{tabular}
\end{adjustbox}
\caption{Performance comparison of various deep learning models on time series forecasting tasks for weather and air pollution using MSE and MAE metrics.}
\label{tab:5}
\end{table*}

\begin{figure}[!htbp]
\centering
\includegraphics[width=0.9\textwidth]{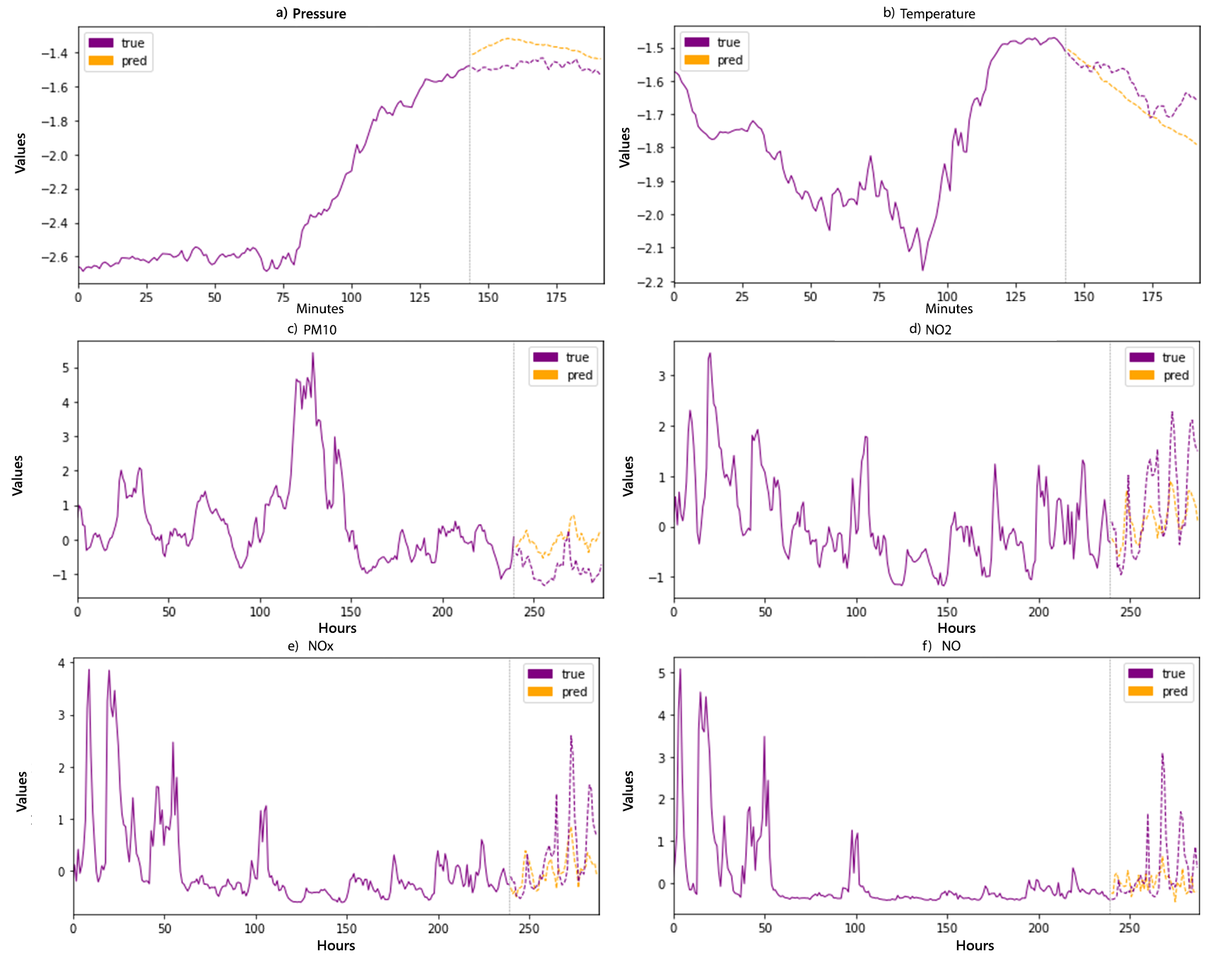}
\caption{The forecasting result of PatchRWKV on Weather (a and b) and Salford Air Pollution dataset (c,d,e and f).}
\label{fig:8}
\end{figure}

Table~\ref{tab:5} presents the performance of various time series processing model on two forecasting tasks. The models are categorised into four types: MLP-based, CNN-based, Transformer-based, and RNN-based. Performance was evaluated using two metrics: Mean Squared Error (MSE) and Mean Absolute Error (MAE). The PatchRWKV approach demonstrates superior performance on both datasets, achieving the lowest MSE and highly competitive MAE values. {Figure~\ref{fig:8}} showed the forecasting results of PatchRWKV on the two real datasets forecasting tasks. The model results were successful in predicting the trends of the pollution indicators. This indicates that PatchRWKV is highly effective in terms of both accuracy and robustness compared to other methods evaluated. Meanwhile, the RNN-based models generally outperformed other types in weather forecasting, with PatchRWKV and LSTM showing particularly strong results.

\subsubsection{Computing Efficiency Analysis}

\begin{table*}[!htbp]
\begin{adjustbox}{max width=\textwidth}
\begin{tabular}{cccccc}
\hline
Method     & Time   Complexity        & Test Step & Parameter  \\ \hline
GRU        & O(L)                     & L         & 38,709,216 \\
LSTM       & O(L)                     & L         & 9,216,192  \\
LightTS    & O(L)                     & 1         & 11,128     \\
DLinear    & O(L)                     & 1         & 18,624     \\
TimesNet   & O(k\textasciicircum{}2L) & 1         & 1,193,781  \\
PatchTST   & O(L\textasciicircum{}2)  & 1         & 677,984    \\
Autoformer & O(LlogL)                 & 1         & 882,709    \\
PatchRWKV  & O(L)                     & 1         & 468,144    \\ \hline
\end{tabular}
\end{adjustbox}
\caption{Computational complexity comparison. L is the sequence length; k is the kernel size of convolutions. The GRU and LSTM are sequential processing that have L test step to capture long-term dependencies. For the rest models, only 1 test step needed for processing the entire sequence in parallel, enabling faster inference.}
\label{tab:6}
\end{table*}

We have conducted a complexity analysis of the PatchRWKV, and its competing models as shown in Table~\ref{tab:6}. GRU, LSTM, LightTS, DLinear, PatchRWKV have a linear time complexity relative to the sequence length L. This is efficient for longer sequences as the complexity grows at a controlled rate. TimesNet is a CNN based model that introduces a quadratic factor based on the kernel size k. PatchTST and Autoformer which are two transformer-based model that have quadratic and logarithmic complexity concerning the sequence length L. For the Test Step, GRU and LSTM are traditional RNN models with L number of test steps, which means they require multiple steps proportional to the length of the sequence, which may increase the computational load during testing or inference. In contrast, models like LightTS, DLinear, TimesNet, PatchTST, Autoformer, and PatchRWKV need only 1 test step as they process the entire sequence in parallel, enabling faster inference. For the count of parameters, GRU (38,709,216) and LSTM (9,216,192) have significantly higher parameter counts compared to others due~to the RNN\textquotesingle s computational mechanism. The proposed PatchRWKV (468,144) have relatively low parameter counts, which generally translates to more lightweight models that are faster and require less memory, making them suitable for scenarios where computational resources are limited.

\subsection{Feature Importance and Model Interpretability}

\begin{figure}[!htbp]
\centering
\includegraphics[width=0.9\textwidth]{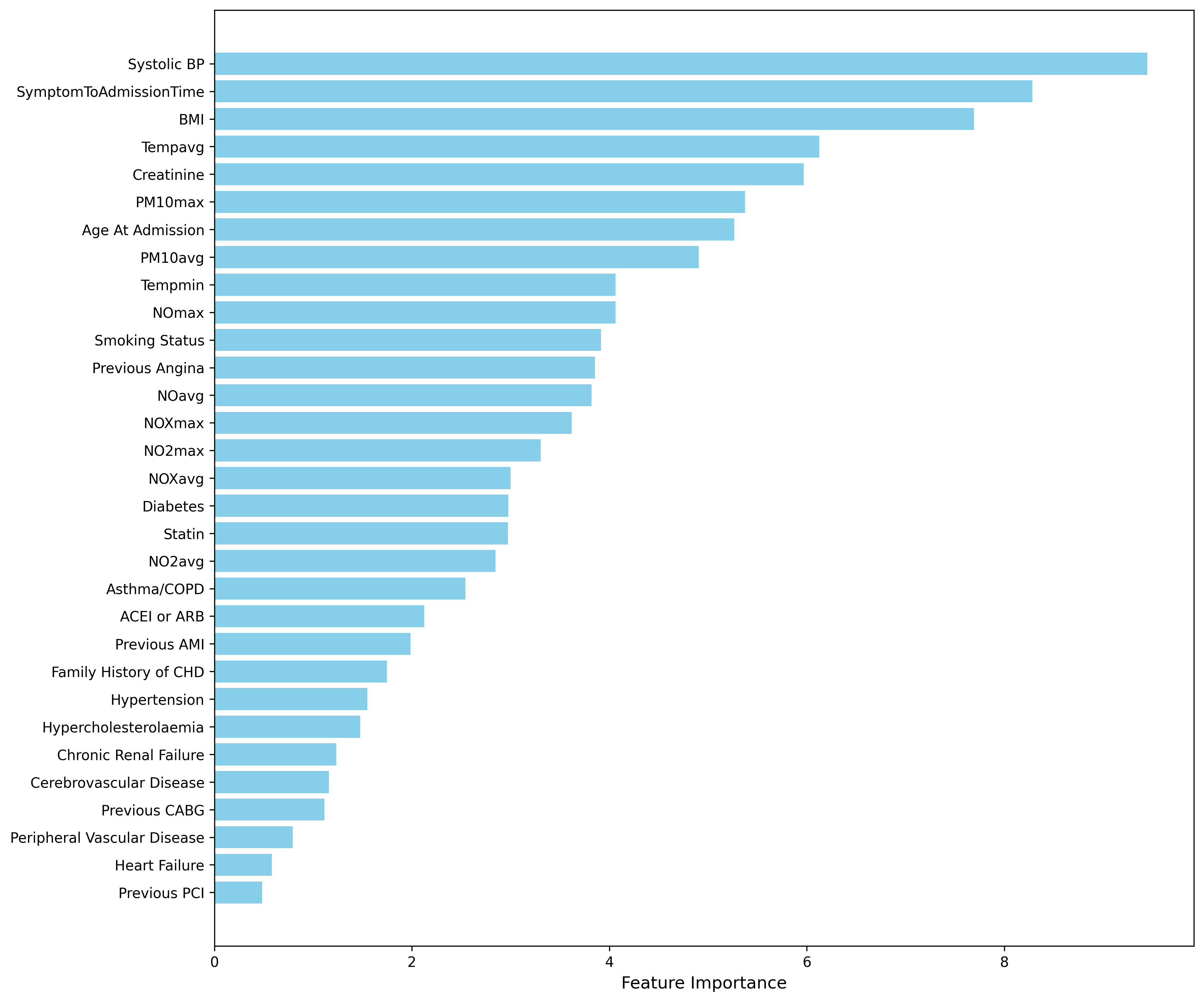}
\caption{Feature importance for ACS subtype prediction (STEMI vs. NSTEMI). Features include clinical, environmental, and historical variables.}
\label{fig:9}
\end{figure}

In this work, we reported feature importance to explain how models make decisions to predict STEMI VS. NSTEMI.The analysis highlights the contributions of 31 risk factors ({Figure~\ref{fig:9}}), encompassing patient demographics, clinical history, vital signs, medications, diagnostic indicators, and environmental factors.

The analysis revealed that Systolic Blood Pressure (SBP) was the most influential predictor (importance score: 9.45), followed closely by Symptom-to-Admission Time (8.29) and Body Mass Index (BMI) (7.69), emphasising the critical roles of cardiovascular function, presentation timing, and metabolic factors. Environmental and physiological markers, such as average temperature (Tempavg) (6.13) and Creatinine levels (5.97), along with air quality metrics like maximum PM$_{10}$ (PM10max) (5.37), were also significant contributors. Moderately important features included Age at Admission (5.26), average PM$_{10}$ (PM10avg) (4.91), minimum temperature (Tempmin) (4.06), maximum NO (NOmax) (4.06), Smoking Status (3.92), and Previous Angina (3.85). Factors like Diabetes (2.98), Statin use (2.97), and Hypertension (1.55) showed relevance but had lower importance scores, suggesting a more indirect role in differentiating STEMI vs. NSTEMI in this model.

\begin{figure}[!htbp]
\centering
\includegraphics[width=0.9\textwidth]{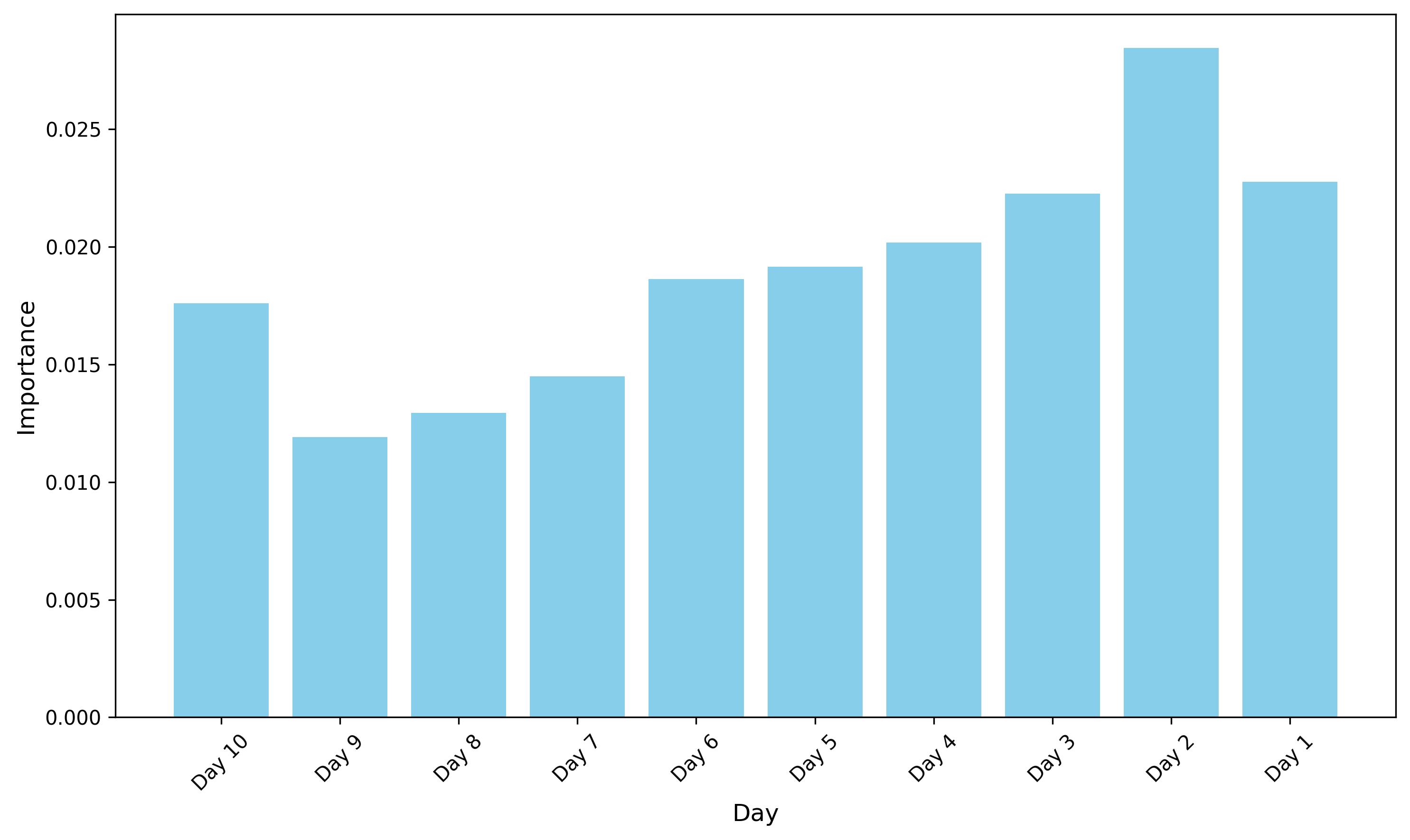}
\caption{Feature Importance by Day for Air Pollution Exposure in ACS Prediction.}
\label{fig:10}
\end{figure}

{Figure~\ref{fig:10}} showed the average importance of air quality each day for predicting ACS risk. The daily step importance analysis revealed that pollution levels on the second day prior to clinical presentation (Day 2) held the highest predictive significance (importance score: 0.0285), closely followed by the day immediately prior (Day 1, importance score: 0.0228) and the third day prior (Day 3, importance score: 0.0223). There was a general trend of declining relevance for days further preceding the event, with Day 9 having the lowest importance (0.0119).
Notably, Day 10 exhibited slightly higher importance (0.0176) than Day 9. This finding highlights the acute impact of short-term pollution exposure on the presentation of ACS, while also suggesting potential influences from earlier exposures or possible modelling artefacts. 

\begin{table*}[!htbp]
    \begin{adjustbox}{max width=\textwidth}
    \begin{tabular}{lrrrrrrl}
    \hline
                          & \multicolumn{3}{c}{\textbf{STEMI}} & \multicolumn{3}{c}{\textbf{NSTEMI}} & \textbf{T-test p-value}            \\ \hline
                          & 25\%          & 50\%          & 75\%          & 25\%          & 50\%          & 75\%          &                            \\ \hline
    \textbf{Systolic BP}    & 111.25        & 127.50        & 146.00        & 120.00        & 136.50        & 154.00        & 0.0019 (\textless 0.05)    \\
    \textbf{SymptomToAdmissionTime} & 1.02          & 5.37          & 20.33         & -213.04       & 6.03          & 33.74         & 0.0020 (\textless 0.05)    \\
    \textbf{BMI}            & 23.32         & 26.04         & 29.89         & 24.52         & 28.03         & 32.23         & 0.0021 (\textless 0.05)    \\
    \textbf{Tempavg}        & 7.60          & 11.26         & 15.17         & 7.12          & 10.61         & 15.03         & 0.0039 (\textless 0.05)    \\
    \textbf{Creatinine}     & 72.25         & 87.00         & 100.75        & 72.00         & 87.00         & 107.00        & 0.0049 (\textless 0.05)    \\
    \textbf{PM10max}        & 22.33         & 28.54         & 41.25         & 22.42         & 30.65         & 44.86         & 0.0487 (\textless 0.05)    \\
    \textbf{Age At Admission}& 59.25         & 71.00         & 81.00         & 59.00         & 72.00         & 81.00         & 0.4911 (\textgreater 0.05) \\
    \textbf{PM10avg}        & 10.61         & 12.94         & 17.04         & 10.62         & 13.68         & 19.24         & 0.0021 (\textless 0.05)    \\
    \textbf{Tempmin}        & 2.20          & 6.20          & 9.20          & 2.20          & 5.70          & 9.20          & 0.5147 (\textgreater 0.05) \\
    \textbf{NOavg}          & 2.44          & 4.73          & 10.01         & 2.35          & 4.23          & 10.29         & 0.0487 (\textless 0.05)    \\
    \textbf{NOmax}          & 11.86         & 33.46         & 80.17         & 12.02         & 29.70         & 87.35         & 0.9342 (\textgreater 0.05) \\
    \textbf{NO2max}         & 39.67         & 54.10         & 66.38         & 41.09         & 53.75         & 69.48         & 0.8793 (\textgreater 0.05) \\
    \textbf{NOXavg}         & 19.31         & 28.95         & 43.00         & 19.67         & 27.09         & 43.95         & 0.628 (\textgreater 0.05)  \\
    \textbf{NO2avg}         & 15.47         & 20.80         & 28.03         & 15.63         & 20.54         & 28.49         & 0.4883 (\textgreater 0.05) \\
    \textbf{NOXmax}         & 56.09         & 100.14        & 177.60        & 57.92         & 94.67         & 196.00        & 0.8761 (\textgreater 0.05) \\ \hline
    \end{tabular}
    \end{adjustbox}
\caption{Comparison of key clinical and environmental predictors for STEMI and NSTEMI cases. The Quartile Cutoff values represent the 25th, 50th (median), and 75th percentiles of each predictor's distribution. The statistical significance of these variables (p $<$ 0.05) supports their clinical relevance in distinguishing STEMI from NSTEMI.}
\label{tab:7}
\end{table*}

Table~\ref{tab:7} summarizes quartile values and t-test results comparing key predictors between STEMI and NSTEMI cases. Significant differences (p \textless 0.05) were observed for several factors. NSTEMI patients showed significantly higher Systolic Blood Pressure, BMI, Creatinine levels, maximum PM$_{10}$, average PM$_{10}$, and average NO compared to STEMI patients. Symptom-to-Admission Time and average temperature also differed significantly between the groups.

Conversely, no significant differences (p \textgreater 0.05) were found for Age at Admission, minimum temperature, or several nitrogen oxide metrics (NOmax, NO2max, NOXavg, NO2avg, NOXmax). These findings suggest NSTEMI presentation in this cohort is associated more strongly with markers of chronic conditions (hypertension, obesity), elevated creatinine, and higher exposure to PM$_{10}$ and average NO.

\section{Discussion}

This study presents TabulaTime, a novel deep learning-based method designed to integrate air pollution data with clinical risk factors for predicting Acute Coronary Syndrome (ACS). The results from our experiments suggest that combining environmental and clinical data significantly improves prediction accuracy, reinforcing the need to include environmental factors in cardiovascular risk models.

\subsection{Model Performance, Innovations, and Clinical Insights}

Traditional cardiovascular risk scores, such as QRISK3, FRS, and GRACE, primarily rely on clinical data and often ignore environmental influences like air pollution. TabulaTime addresses this gap by integrating time-series air pollution data with clinical factors, providing a more comprehensive risk assessment framework. This integration is crucial, given the substantial evidence linking pollutants such as PM$_{10}$, NO$_{2}$, NO, and NO$_{x}$ to adverse cardiovascular outcomes via mechanisms like inflammation and oxidative stress \cite{Chen2022Hourly,Braunwald2023Air}. Our feature importance analysis ({Figure~\ref{fig:9}}) supports this, identifying air quality metrics, specifically PM$_{10}$ (maximum and average) and average NO, as significant contributors alongside key clinical factors in differentiating ACS presentations. Crucially, our experiments demonstrate that incorporating air pollution data significantly enhances predictive performance, improving the accuracy of TabulaTime by 10.1\% compared to using clinical data alone. This finding aligns with epidemiological studies linking short-term air pollutant exposure (including PM$_{10}$ and NO$_{x}$) with increased cardiovascular events. By leveraging real-time environmental exposure data, TabulaTime offers a more holistic and timely assessment of ACS risk, capturing the immediate impact of environmental triggers.


One of the major innovations in TabulaTime is the PatchRWKV module, an efficient time-series feature extraction module that combines Recurrent Neural Networks (RNNs) and Transformer-based self-attention mechanisms. Our experiments across multiple classification and forecasting tasks confirmed that PatchRWKV surpassed state-of-the-art models such as PatchTST, Autoformer, and TimesNet, while maintaining linear computational complexity. The patching strategy, which isolates local temporal patterns (e.g., daily pollution cycles), allows the model to discern subtle exposure-response relationships that manual feature engineering might miss. This capability is critical for public health applications, where identifying high-risk periods (e.g., days with PM$_{10}$ \textgreater 40 µg/m³) can inform real-time interventions.

When compared to Random Forest (RF), LightGBM, and CatBoost, TabulaTime significantly outperformed these traditional machine learning models, achieving a 32.2\% improvement over RF and a 20.5\% improvement over CatBoost. Notably, even without environmental data, TabulaTime still maintains a superior performance, suggesting that its novel multimodal feature integration and advanced temporal feature extraction contribute significantly to its predictive strength.

The robustness of TabulaTime was tested using data from the COVID-19 lockdown period (March 2020–December 2021), when air pollution levels were significantly lower than usual. Despite these different environmental conditions compared to the pre-lockdown training data, the model maintained a high predictive accuracy of 78.8\%. This indicates that TabulaTime is adaptable and not solely reliant on the specific pollution patterns present during training. It likely adjusted its predictions, possibly by giving more weight to clinical data when the usual pollution signals were weaker. This ability to handle changing environmental factors highlights the model's suitability for real-world deployment, such as in health monitoring systems where conditions can vary.

The interpretability features of TabulaTime, derived from attention mechanisms and feature importance analysis, provide valuable clinical insights. The analysis ({Figure~\ref{fig:9}}) highlights that established clinical risk factors like Systolic Blood Pressure (SBP), Symptom-to-Admission Time, and Body Mass Index (BMI) are primary drivers in differentiating STEMI from NSTEMI presentations. This aligns with clinical understanding where factors reflecting hemodynamic stability, presentation delay, and metabolic health significantly influence ACS type and severity. Notably, the substantial importance assigned to environmental factors, particularly maximum and average PM$_{10}$ and average NO, alongside clinical markers like Creatinine and average temperature, underscores a complex interplay between acute environmental triggers, underlying physiological state (e.g., renal function), and the manifestation of specific ACS types. This finding strongly advocates for the integration of environmental monitoring into routine cardiovascular risk assessment and public health strategies.

The temporal analysis ({Figure~\ref{fig:10}}) highlights a critical exposure window, with pollution levels 1-3 days prior (peaking at Day 2) showing the highest importance. This strongly indicates that short-term, recent exposure is a significant trigger for acute cardiovascular events, underscoring the clinical and public health value of timely air quality alerts and interventions during high-pollution periods. 

The analysis also revealed a slight, though notable, increase in importance for Day 10 compared to Day 9. From a modelling standpoint, in sequential models such as PatchRWKV, earlier time steps in the input sequence (e.g., Day 10) may receive relatively stronger attention weights due to their positional encoding or sequence alignment, potentially influencing feature importance scores. This effect can sometimes exaggerate the influence of positions within fixed-length temporal windows. From a clinical perspective, the finding may reflect a lagged biological response to air pollution exposure. Prior studies have reported delayed effects of environmental pollutants on vascular inflammation, endothelial dysfunction, and thrombotic pathways \cite{Kuzma2021Impact,robertson2018ambient,al2020environmental}, which may increase vulnerability to ACS several days after exposure. Furthermore, histopathological evidence suggests that plaque erosion—a mechanism responsible for ~25\% of ACS cases—may initiate approximately 7 days prior to clinical presentation, as evidenced by remodelling of the thrombus overlying the disrupted plaque \cite{white2016endothelial,kramer2010relationship}. It is tempting to speculate that the increased peak around 10 days prior to clinical presentation may represent a plaque erosion population, however this would require further evidence to support such a hypothesis. Smoking is the only known modifiable risk factor for plaque erosion, which could potentially be regarded as the most severe ‘air pollution’ model available.

\subsection{Limitations and Future Work}


The TabulaTime framework is a flexible and extensible multimodal learning model that integrates both tabular and time-series data for acute coronary syndrome (ACS) subtype prediction. Its modular design enables the incorporation of diverse clinical and environmental inputs, as demonstrated by the seamless addition of external temperature data. TabulaTime outperformed traditional models and effectively captured short-term pollution effects, offering new insights into environmental and physiological contributions to ACS risk.

While TabulaTime has demonstrated strong performance and clinical relevance, there remain several opportunities for further enhancement.

First, the current model was developed using data from a single air quality monitoring station in Salford (UKA00339), which limited the inclusion of some environmental variables. Key meteorological factors such as humidity, atmospheric pressure, rainfall, and wind direction—known to influence pollution dynamics and cardiovascular outcomes—were not available at the time of analysis. Nevertheless, the successful integration of external temperature data illustrates the model’s adaptability, and other climatic variables will be considered in future iterations. Geographic and collective factors such as traffic density and high-resolution socioeconomic status (SES) indicators were also not included due to data unavailability, but they represent valuable additions to improve spatial granularity and context-awareness.

Second, the model was evaluated in a single urban setting. Although Salford represents a socioeconomically diverse population (mean Townsend score ~4), broader validation is needed to ensure generalisability across other regions. Encouragingly, the planned expansion to Greater Manchester will support the inclusion of spatially resolved environmental and demographic data, enhancing the model’s robustness and applicability in diverse real-world settings.

Third, although cardiovascular risk scores such as QRISK and the Framingham Risk Score (FRS) are referenced, no direct comparisons were made, as these tools are designed for long-term population-level prediction. In contrast, TabulaTime addresses a specific classification task among patients already diagnosed with ACS. Nevertheless, its architecture offers potential for extension to broader population-level risk prediction in future studies.

Fourth, a major consideration in the current study is the class imbalance between STEMI (n = 123) and NSTEMI (n = 587) cases. Conducting subgroup analyses on this already skewed dataset, particularly within the smaller STEMI group, would result in very small sample sizes per stratum. This introduces a high risk of overfitting, statistical noise, and unreliable effect estimates, which would compromise the validity of any conclusions drawn from such analyses. We believe that performing subgroup analysis under these conditions could be misleading rather than informative. Therefore, we have chosen not to report subgroup-specific performance in the current study, in order to avoid drawing potentially unstable or spurious conclusions. However, we recognise the importance of subgroup insights and are actively planning targeted subgroup analyses in future work, supported by expanded datasets to ensure more robust and meaningful conclusions.

Future work will focus on expanding the model’s geographic scope and input feature set. This includes incorporating additional meteorological variables (e.g., wind direction, pressure, rainfall), as well as spatial and socioeconomic factors such as traffic density and area-level deprivation metrics. The model will be validated across the Greater Manchester region using more comprehensive and spatially resolved datasets. As we expand the cohort to include a larger and more balanced population, we intend to perform a comprehensive investigation of model performance across clinically meaningful subgroups (e.g., by age, cardiovascular history, and risk factor burden), ensuring fairness and generalisability. Such work will be essential for supporting personalised deployment of the TabulaTime model in real-world settings. Furthermore, we explicitly state our intention to apply TabulaTime to multimodal clinical time series datasets, such as those found in MIMIC-IV, eICU, or wearable health platforms. These data sources offer ideal testbeds to demonstrate how TabulaTime can support applications such as early deterioration prediction, dynamic patient risk profiling, and real-time decision support. Finally, as more population-based data become available, TabulaTime may be adapted for primary prevention use cases and directly compared with traditional cardiovascular risk scoring tools for incident ACS prediction.

\section{Conclusion}


This study introduced TabulaTime, a novel multimodal deep learning framework designed to integrate time-series environmental data (air pollution and climate) with clinical tabular data for enhanced prediction of Acute Coronary Syndrome (ACS) presentation (STEMI vs. NSTEMI). TabulaTime effectively leverages the strengths of both data modalities, offering a more comprehensive approach to cardiovascular risk assessment than traditional clinical-only models. A key innovation is the PatchRWKV module, which provides efficient and accurate automatic feature extraction from time-series data, capturing complex temporal patterns with linear computational complexity.

Our experimental evaluation demonstrated TabulaTime's superior performance. It significantly outperformed established machine learning models (Random Forest, LightGBM, CatBoost) by 20.5\% to 32.2\% in accuracy for differentiating ACS subtypes. Crucially, the integration of environmental data improved TabulaTime's accuracy by 10.1\%, highlighting the substantial contribution of air pollution and climate factors to predictive modelling. The PatchRWKV module also excelled independently, surpassing state-of-the-art models in benchmark time-series classification and forecasting tasks. Furthermore, TabulaTime showed robustness, maintaining high accuracy even during the COVID-19 lockdown period despite significant changes in air pollution levels. Feature importance analysis confirmed the relevance of both clinical factors (e.g., Systolic Blood Pressure, Symptom-to-Admission Time, BMI) and environmental variables (e.g., PM$_{10}$, NO, temperature), particularly emphasizing the impact of short-term pollution exposure in the days preceding an ACS event.

Future work will focus on expanding the model's geographic scope, incorporating a richer set of environmental and socioeconomic features, and validating its performance across diverse populations and larger, more balanced datasets. Planned subgroup analyses will explore performance variations, and the framework will be applied to other complex clinical time-series datasets (e.g., MIMIC-IV, eICU) and potentially adapted for primary ACS prevention.

In conclusion, TabulaTime represents a significant advancement in integrating environmental health data with clinical risk factors. Its accuracy, efficiency, and interpretability offer considerable potential for improving clinical decision support, informing public health strategies, and advancing personalised cardiovascular medicine.

\section*{Acknowledgement}

Funding support: British Heart Foundation Doug Gurr Cardiovascular Catalyst Award (BHF-CC/22/250021). Liangxiu Han is supported by EPSRC (EP/X013707/1).


\bibliography{sn-bibliography}

\end{document}